\documentclass{article}

\usepackage{microtype}
\usepackage{graphicx}
\usepackage{subfigure}
\usepackage{booktabs} 

\usepackage{hyperref}

\usepackage[accepted]{icml2024}

\usepackage{amsmath}
\usepackage{amssymb}
\usepackage{mathtools}
\usepackage{amsthm}
\usepackage[acronyms,shortcuts]{glossaries} 

\usepackage[capitalize,noabbrev]{cleveref}

\theoremstyle{plain}

\theoremstyle{definition}

\theoremstyle{remark}

\usepackage[textsize=tiny]{todonotes}

\icmltitlerunning{Time Weaver: A Conditional Time Series Generation Model}

\begin{document}

\twocolumn[
\icmltitle{Time Weaver: A Conditional Time Series Generation Model}




\begin{icmlauthorlist}
\icmlauthor{Sai Shankar Narasimhan}{yyy}
\icmlauthor{Shubhankar Agarwal}{yyy}
\icmlauthor{Oguzhan Akcin}{yyy}
\icmlauthor{Sujay Sanghavi}{yyy}
\icmlauthor{Sandeep Chinchali}{yyy}
\end{icmlauthorlist}

\icmlaffiliation{yyy}{Department of Electrical and Computer Engineering, The University of Texas at Austin, Austin, USA}

\icmlcorrespondingauthor{Sai Shankar Narasimhan}{nsaishankar@utexas.edu}

\icmlkeywords{Time series Generation, Diffusion Models, Conditional Generation Evaluation}

\vskip 0.3in
]



\printAffiliationsAndNotice{} 
\crefname{problem}{Problem}{Problems}
\crefname{assumption}{Assumption}{Assumptions}
\crefname{remark}{Remark}{Remarks}
\crefname{definition}{Defn.}{Defns.}
\crefname{proposition}{Prop.}{Props.}
\crefname{algorithm}{Alg.}{Algs.}
\crefname{equation}{Eq.}{Eqs.}
\crefname{figure}{Fig.}{Figs.}
\crefname{section}{Sec.}{Secs.}
\crefname{appendix}{App.}{App.}

\newcommand{\timeweaver}{\textsc{Time Weaver}}
\newcommand{\jointrealdistribution}{D_r^{\textnormal{emb}}}
\newcommand{\jointgenerateddistribution}{D_g^{\textnormal{emb}}}
\newcommand{\distribution}{\mathcal{D}}
\newcommand{\dataset}{D}
\newcommand{\timeseriesdataset}{X}
\newcommand{\timeseriesdatadistribution}{\mathcal{X}}
\newcommand{\metadataset}{C}
\newcommand{\conditionalembeddings}{z}
\newcommand{\categoricalembeddings}{z_{\textnormal{cat}}}
\newcommand{\continuousembeddings}{z_{\textnormal{cont}}}
\newcommand{\timeseries}{x}
\newcommand{\timetotimeloss}{\loss^{x \rightarrow x}}
\newcommand{\timetocondloss}{\loss^{x \rightarrow c}}
\newcommand{\condtocondloss}{\loss^{c \rightarrow c}}
\newcommand{\similarity}{d}

\newcommand{\jointdatadistribution}{p(\timeseries,\condition)}
\newcommand{\jointdataset}{D_{x,c}}
\newcommand{\conditionaldatadistribution}{p(\timeseries | \condition)}
\newcommand{\generationfunction}{G}

\newcommand{\nchannels}{F}
\newcommand{\timeserieslength}{L}
\newcommand{\nconditions}{K}
\newcommand{\timeseriesdomain}{\mathbb{R}^{\timeserieslength \times \nchannels}}
\newcommand{\conditiondomain}{\mathbb{R}^{\timeserieslength \times \nconditions}}
\newcommand{\catconditiondomain}{\mathbb{R}^{\timeserieslength \times \nconditions_{\textnormal{cat}}}}
\newcommand{\contconditiondomain}{\mathbb{R}^{\timeserieslength \times \nconditions_{\textnormal{cont}}}}
\newcommand{\cattokendomain}{\mathbb{R}^{\timeserieslength \times d_{\textnormal{cat}}}}
\newcommand{\conttokendomain}{\mathbb{R}^{\timeserieslength \times d_{\textnormal{cont}}}}
\newcommand{\zdomain}{\mathbb{R}^{\timeserieslength \times d_{\textnormal{meta}}}}

\newcommand{\generativemodel}{f}
\newcommand{\cattoken}{\theta_{\textnormal{token}}^{\textnormal{cat}}}
\newcommand{\conttoken}{\theta_{\textnormal{token}}^{\textnormal{cont}}}
\newcommand{\salayer}{\theta_{\textnormal{condn}}}
\newcommand{\loss}{\mathcal{L}}
\newcommand{\totalloss}{\loss}
\newcommand{\diffusionloss}

\newcommand{\denoiser}{\mathcal{\theta_{\textnormal{denoiser}}}}
\newcommand{\conditionembedder}{\theta_{\textnormal{condn}}}
\newcommand{\metricembedderts}{\phi_{\textnormal{time}}}
\newcommand{\metricembeddercondn}{\phi_{\textnormal{meta}}}

\newcommand{\batchsize}{N_{\textnormal{batch}}}
\newcommand{\patchsize}{N_{\textnormal{patch}}}
\newcommand{\patchlength}{L_{\textnormal{patch}}}

\newcommand{\condition}{c}
\newcommand{\conditioncategorical}{c_{\textnormal{cat}}}
\newcommand{\conditioncontinuous}{c_{\textnormal{cont}}}
\newcommand{\conditioncategoricaldomain}{\mathbb{N}^{L \times K_{\textnormal{cat}}}}
\newcommand{\conditioncontinuousdomain}{\mathbb{R}^{L \times K_{\textnormal{cont}}}}
\newcommand{\numcategoricalfeatures}{K_{\textnormal{cat}}}
\newcommand{\numcontinuousfeatures}{K_{\textnormal{cont}}}

\glsdisablehyper
\renewcommand{\acrfullformat}[2]{#1~(#2)} 
\newacronym{gan}{GAN}{Generative Adversarial Network}
\newacronym{ecg}{ECG}{electrocardiogram}
\newacronym{fid}{FID}{Frechet Inception Distance}
\newacronym{jftsd}{J-FTSD}{Joint Frechet Time Series Distance}
\newacronym{dm}{DM}{Diffusion Model}
\newacronym{tstr}{TSTR}{train on synthetic test on real}
\newacronym{fd}{FD}{Frechet Distance}
\newacronym{fjd}{FJD}{Frechet Joint Distance}
\newacronym{cnn}{CNN}{Convolutional Neural Network}
\newacronym{auc}{AUC}{area under the curve}
\newacronym{fc}{FC}{fully connected}


\newcommand{\oguzhan}[1]{{\color{blue} #1}}
\newcommand{\oguzhanc}[2]{{\color{red}\sout{#1}}{\color{blue}#2}}
\newcommand{\sai}[1]{{\color{cyan} [Sai]: #1}}
\newcommand{\somi}[1]{{\color{green} [Somi]:#1}}
\newcommand{\sandeep}[1]{{\color{magenta}[Sandeep]: #1}}
\newcommand{\changes}[1]{{\color{blue} [Edits]: #1}}

\renewcommand{\thefootnote}{\fnsymbol{footnote}}
\begin{abstract}
Imagine generating a city's electricity demand pattern based on weather, the presence of an electric vehicle, and location, which could be used for capacity planning during a winter freeze. Such real-world time series are often enriched with paired heterogeneous contextual metadata (e.g., weather and location). Current approaches to time series generation often ignore this paired metadata. Additionally, the heterogeneity in metadata poses several practical challenges in adapting existing conditional generation approaches from the image, audio, and video domains to the time series domain. To address this gap, we introduce \timeweaver, a novel diffusion-based model that leverages the heterogeneous metadata in the form of categorical, continuous, and even time-variant variables to significantly improve time series generation. Additionally, we show that naive extensions of standard evaluation metrics from the image to the time series domain are insufficient. These metrics do not penalize conditional generation approaches for their poor specificity in reproducing the metadata-specific features in the generated time series. Thus, we innovate a novel evaluation metric that accurately captures the specificity of conditional generation and the realism of the generated time series. We show that \timeweaver \space outperforms state-of-the-art benchmarks, such as \acp{gan}, by up to 30\% in downstream classification tasks on real-world energy, medical, air quality, and traffic datasets. Code\footnote{https://github.com/UTAustin-SwarmLab/TimeWeaver-CPS/tree/main} and checkpoints are publicly available.
\end{abstract}
\renewcommand{\thefootnote}{\arabic{footnote}}

\section{Introduction}
\label{sec:introduction}
Generating synthetic time series data is useful for creating realistic variants of private data \cite{yoon2020anonym}, stress-testing production systems with new scenarios \cite{rizza2022stress, agarwal2022task}, asking “what-if” questions, and even augmenting imbalanced datasets \cite{gowal2021improving}. Imagine generating a realistic medical \ac{ecg} pattern based on a patient’s age, gender, weight, medical record, and even the presence of a pacemaker. This generated data could be used to train medical residents, sell realistic data to third parties (anonymization), or even stress-test a pacemaker’s ability to detect diseases on rare variations of \ac{ecg} data.
\begin{figure}[!t]
    \centering
    \includegraphics[width=\columnwidth]{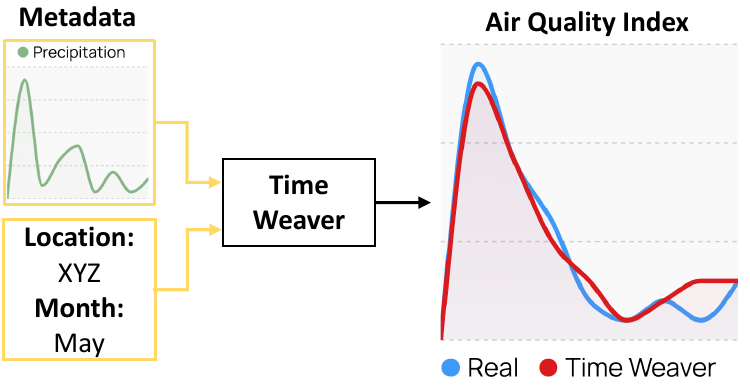}
    \vspace{-2em}
    \caption{{\textbf{\timeweaver \space generates realistic metadata-specific time series.} Consider generating the air quality index of a particular location (\texttt{XYZ}) given the expected precipitation (green) for a specific month (\texttt{May}). \timeweaver \space uses these metadata features to generate samples (red) that closely match reality (blue). }}
    \label{fig:motivation}
    \vspace{-1em}
\end{figure}

Despite potential advantages, current time series generation methods \cite{yoon2019TimeGAN, jeha2021psa, donahue2019wavegan} ignore the rich contextual metadata and are incapable of generating time series for specific real-world conditions. This is not due to a lack of data, as standard time series datasets have long come with paired metadata conditions. Instead, it is because today's methods cannot handle diverse metadata conditions. 

\begin{figure}[!t]
    \centering
    \includegraphics[width=0.4\textwidth, keepaspectratio]{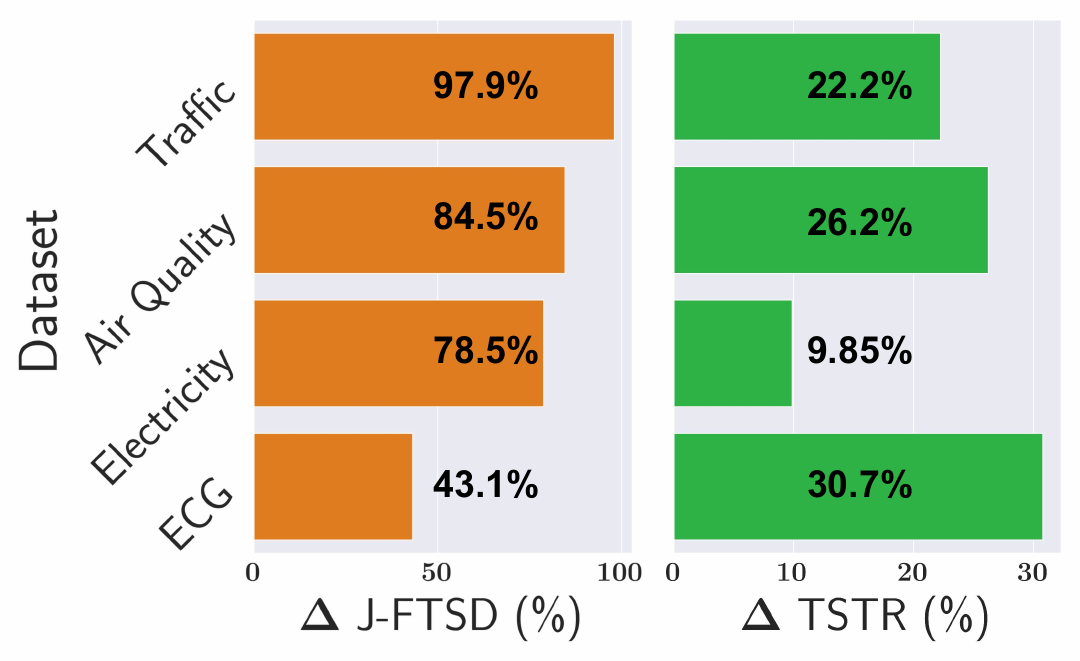}
    \caption{\textbf{\timeweaver \space beats \ac{gan}s on all datasets on the Joint Frechet Time Series Distance (\ac{jftsd}) and Train on Synthetic Test on Real (TSTR) metrics.} \ac{jftsd} indicates the distributional similarity between the generated and real time series datasets. Lower values of \ac{jftsd} indicate that both generated and real time series distributions are closer. TSTR indicates the performance of a downstream task model trained on generated time series data and evaluated on real time series data. Higher values of TSTR indicate higher quality of the generated time series data. We show the percentage improvement of \timeweaver \space over state-of-the-art \ac{gan} models on four diverse datasets.}
    \label{fig:motivation_metrics}
    \vspace{-1.25em}
\end{figure}

At first glance, generating realistic time series based on rich metadata conditions might seem like a straightforward extension of conditional image, video, or audio generation \cite{rombach2021highresolution, ramesh2022hierarchical, kong2021diffwave}. However, we argue that there are practical differences that make conditional time series generation and evaluation challenging, which are:
\begin{enumerate}
    \item \textbf{Rich Metadata:} Metadata can be categorical (e.g., whether a patient has a pacemaker), quantitative (e.g., age), or even a time series, such as anticipated precipitation. Any conditional generative model for time series should incorporate such a diverse mix of metadata conditions (\cref{tab:dataset}). In contrast, image, video, and audio generation often deal with static text prompts. 
    \item \textbf{Visual Inspection of Synthetic Data Quality:} Visual inspection is a key aspect in evaluating image generation approaches as evaluation metrics like the Inception Score (IS) are widely adopted due to their alignment with human judgment. On the contrary, it is non-trivial to glance at a time series and tell if it retains key features, such as statistical moments or frequency spectra. 
    \item \textbf{Architectural Differences:} In the image and audio domains, we have powerful feature extractors trained on internet-scale data \cite{radford2021learning, laionclap2023}. These are vital building blocks for encoding conditions in image generation \cite{rombach2021highresolution}. However, these models are non-existent in the time series domain due to the irregular nature of the time series datasets with respect to horizon lengths, number of channels, and the heterogeneity of the metadata.
    \item \textbf{Evaluation Metrics:} Evaluating conditional generation approaches requires a metric that captures the specificity of the generated samples with respect to their paired metadata. In \cref{fig:metric_comparison}, we show how the existing metrics, such as the time series equivalent of the standard \ac{fid} score \cite{jeha2021psa}, fail to capture this specificity and only measure how close the real and generated data distributions are. This is because these metrics completely ignore the paired metadata in their evaluation.
\end{enumerate}
Given the above differences and insufficiencies in metrics, \textbf{our contributions} are:
\begin{enumerate}
    \item We present \timeweaver \space (\cref{fig:motivation}), a novel diffusion model for generating realistic multivariate time series conditioned on metadata. We specifically innovate on the standard diffusion model architecture to process categorical and continuous metadata conditions.
    \item We propose the \ac{jftsd}, specifically designed to evaluate conditional time series data generation models. \ac{jftsd} incorporates time series and metadata conditions with feature extractors trained using a contrastive learning framework. In \cref{sec:experiments}, we showcase \ac{jftsd}'s ability to accurately rank approaches based on their ability to model conditional time series data distributions.
    \item We show that our approach significantly outperforms the state-of-the-art \ac{gan} models in generating high-quality, metadata-specific time series on real-world energy, healthcare, pollution, and traffic datasets (\cref{fig:motivation_metrics}).   
\end{enumerate}

\section{Background and Related Works}
\label{sec:related_work}
\textbf{Generative Models in Time Series:} Recently, \acfp{gan} \cite{donahue2019wavegan, yoon2019TimeGAN, Li2022TTSGANAT, thambawita2021pulse2pulse} have emerged as popular methods for time series data generation. However, these \ac{gan}-based approaches often struggle with unstable training and mode collapse \cite{chen2021surveygans}. In response, \acp{dm} \cite{sohl2015deep} have been introduced in the time series domain \cite{alcaraz2023diffusion, tashiro2021csdi}, offering more realistic data generation. \acp{dm} are a class of generative models that are state-of-the-art in a variety of domains, including image \cite{dhariwal2021diffusion, ho2020denoising}, speech \cite{chen2020wavegrad,kong2021diffwave}, and video generation \cite{ho2022video}. \acp{dm} operate by defining a Markovian forward process $q$. The forward process gradually adds noise to a clean data sample $x_0 \sim \mathcal{X}$, where $\mathcal{X}$ is the data distribution to be learned. The forward process is predetermined by fixing a noise variance schedule $\left\{ \beta_1, \ldots, \beta_T \right \}$, where $\beta_t \in \left[ 0,1 \right]$ and $T$ is the total number of diffusion steps. The following equations describe the forward process: 
\begin{align}
    &q(x_1, \ldots, x_T \mid x_0)  = \prod_{t=1}^{T} q(x_t \mid x_{t-1}), \\
    &q(x_t \mid x_{t-1}) = \mathcal{N}(\sqrt{1-\beta_t}x_{t-1}, \beta_t \mathbf{I}).
    \label{eq:forward_process}
\end{align}
Here, $\mathcal{N}(\mu, \Sigma)$ represents a Gaussian distribution with mean $\mu$ and covariance matrix $\Sigma$. During training, a clean sample $x_0$ is transformed into $x_t$ using \cref{eq:forward_process}. Then, a neural network, $\denoiser(x_t,t)$, is trained to estimate the amount of noise added between $x_{t-1}$ and $x_t$ with the following loss function:
\begin{equation}
     \loss_{\text{DM}} = \mathbb{E}_{x \sim \mathcal{X}, \epsilon \sim \mathcal{N}(\textbf{0}, \textbf{I}),t \sim \mathcal{U}(1,T)} \left[ \| \epsilon - \denoiser(x_t,t) \right \|^2_2]. \label{eq:dm_objective}
\end{equation}
Here, $t \sim \mathcal{U}(1, T)$ indicates that $t$ is sampled from a uniform distribution between 1 and $T$, and $\epsilon$ is the noise added to $x_{t-1}$ to obtain $x_t$. In inference, we start from $x_T \sim \mathcal{N}(\mathbf{0}, \mathbf{I})$, where $\mathcal{N}(\mathbf{0}, \mathbf{I})$ represents a zero mean, unit variance Gaussian distribution, and iteratively denoise using $\denoiser$ to obtain a clean sample from the data distribution $\mathcal{X}$, \emph{i.e.}, $x_T \rightarrow x_{T-1}, \ldots, x_0$. A detailed explanation of \acp{dm} is provided in \cref{appendix:diffusion_primer}.

For conditional \acp{dm}, the most commonly used approach is to keep the forward process the same as in \cref{eq:forward_process}, and add additional conditions $\condition$ to the reverse process. Minimizing $\left\| \epsilon - \denoiser(x_t,t,c) \right \| ^2_2$ in the loss function provided in \cref{eq:dm_objective} facilitates learning the conditional distribution. Conditional \acp{dm} are used in image, video \cite{saharia2021palette, andreas2022repaint, rombach2021highresolution, ramesh2022hierarchical}, and speech \cite{kong2021diffwave} generation. These models allow for diverse conditioning inputs, like text, image, or even segmentation maps. However, these methods rely on image-focused tools like \acp{cnn}, which struggle to maintain essential time series characteristics such as long-range dependencies, as noted in \cite{gu2022efficiently}. For time series data, models such as CSDI \cite{tashiro2021csdi} and SSSD \cite{alcaraz2022diffusionbased} exist but are mainly limited to imputation tasks without substantial conditioning capabilities. Closest to our work, \citet{alcaraz2023diffusion} attempt to incorporate \ac{ecg} statements as metadata (only categorical) for \ac{ecg} generation. However, this approach falls short as it does not consider heterogeneous metadata. Our method surpasses these limitations by effectively handling a broader range of metadata modalities, thus enabling more realistic time series data generation under varied heterogeneous conditions.

\textbf{ Metrics for Conditional Time Series Generation:} Various metrics have been developed in the time series domain, focusing on the practical utility of the generated time series data. To this end, the Train on Synthetic Test on Real (TSTR) metric \cite{Jordon2018PATEGANGS, esteban2017realvalued} is used to assess the ability of synthetic data to capture key features of the real dataset. TSTR metrics have been widely used to evaluate unconditional time series generation. \citet{yoon2019TimeGAN} proposed the predictive score where synthetic time series data is used to train a forecaster, and the forecaster's performance is evaluated on real time series data. More traditional approaches include average cosine similarity, Jensen distance \cite{Li2022TTSGANAT}, and autocorrelation comparisons \cite{Lin2020Using, bahrpeyma2021timegenval}. However, these heuristics often fail to fully capture the nuanced performance of conditional generative models.

A more popular method to evaluate generative models is to use distance metrics between the generated and real data samples. One of the most commonly used distance metrics is the \ac{fd} \cite{frechet:hal-04093677}. The \ac{fd} between two multivariate Gaussian distributions $\distribution_1 \sim \mathcal{N}(\mu_1, \Sigma_1)$ and $\distribution_2 \sim \mathcal{N}(\mu_2, \Sigma_2)$ is:
\begin{equation}
    FD(\distribution_1, \distribution_2) = \| \mu_1 - \mu_2 \|^2 + \textnormal{Tr}(\Sigma_1 + \Sigma_2 - 2(\Sigma_1 \Sigma_2)^\frac{1}{2}).
\label{eq:fd}
\end{equation}
To evaluate image generation models, the FD is adjusted to the Frechet Inception Distance (\ac{fid}) \cite{heusel2017gans}. \ac{fid} uses a feature extractor, the Inception-v3 model \cite{Szegedy2015RethinkingTI}, to transform images into embeddings upon which the \ac{fd} is calculated. Similar adaptations such as the Frechet Video Distance \cite{unterthiner2018towards}, Frechet ChemNet Distance \cite{preuer2018frechet}, and Context-FID \cite{jeha2021psa} exist for other domains, employing domain-specific feature extractors. However, these metrics are designed only to evaluate unconditional data generation since they only match the true data distribution marginalizing over all the conditions.

To evaluate conditional generation models, many metrics are proposed for categorical conditions \cite{murray2019pfagan, huang2018multimodal, Benny2020EvaluationMF,liu2018improved,miyato2018cgans}. To create a more general metric, \citet{soloveitchik2022conditional} proposed the conditional FID (CFID) metric that works with continuous conditionals and calculates the conditional distributions of the generated and real data given the condition. In particular, \citet{devries2019evaluation} propose the \ac{fjd}, where the embeddings of the image and condition are obtained with different embedding functions and concatenated to create a joint embedding space. \citet{devries2019evaluation} consider conditions that are classes (image category), text descriptions (image captions), or images (for tasks like style transfer). However, in our case, the metadata could be any arbitrary combination of categorical, continuous, and time-varying conditions. Additionally, like other metrics considered in the literature, \ac{fjd} is defined for image generation and does not consider the unique characteristics of time series data. In contrast, our proposed \ac{jftsd} metric is specifically
designed to evaluate time series data generation models
conditioned on heterogeneous metadata.

\section{Problem Formulation}
\label{sec:problem_formulation}
\begin{figure*}[!ht]
\centering
\includegraphics[width=\textwidth]{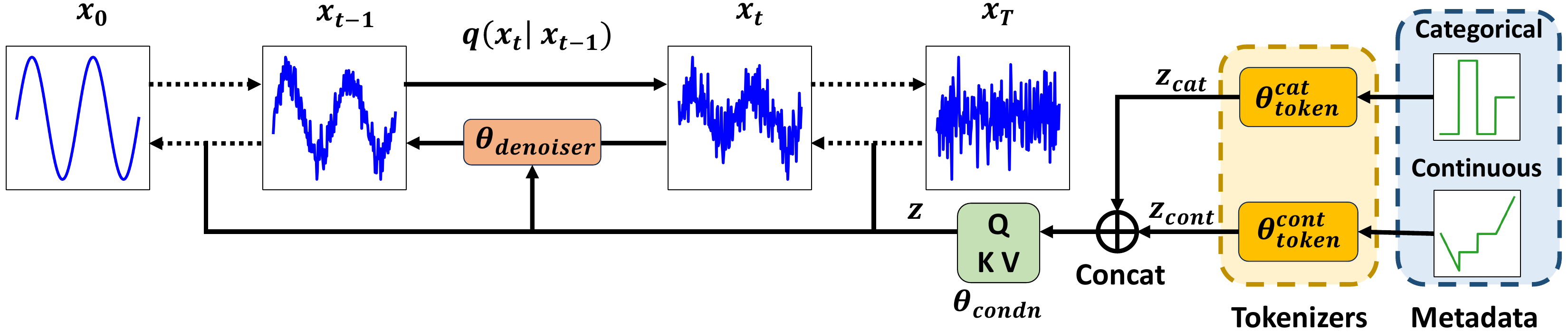}
\vspace{-2em}
\caption{\textbf{\timeweaver \space architecture for incorporating metadata in the diffusion process:} This figure shows the training and inference processes of \timeweaver. For training, we start from the original sample $x_0$ (on the left) and gradually add noise through a forward process $q(x_t \mid x_{t-1})$, resulting in noisy samples $x_t$. The denoiser, $\denoiser$, is trained to estimate the amount of noise added to obtain $x_t$ from $x_{t-1}$. During inference, the categorical and continuous metadata are first preprocessed with tokenizers $\cattoken$ and $\conttoken$, respectively. Then, we concatenate their output and process it through a self-attention layer $\salayer$ to create the metadata embedding $\conditionalembeddings$. This embedding is fed into $\denoiser$ with the noisy sample $x_t$ to obtain $x_{t-1}$. The denoising process is repeated for $T$ diffusion steps to obtain a clean sample similar to $x_0$.}
\vspace{-1em}
\label{fig:arch}
\end{figure*}

Consider a multivariate time series sample $\timeseries \in \timeseriesdomain$, where $\timeserieslength$ denotes the time series horizon, and $\nchannels$ denotes the number of channels. Each sample $\timeseries$ is associated with metadata $\condition$, comprising categorical features $\conditioncategorical \in \conditioncategoricaldomain$ and continuous features $\conditioncontinuous \in \conditioncontinuousdomain$. Here, $\numcategoricalfeatures$ and $\numcontinuousfeatures$ indicate the total numbers of categorical and continuous metadata features, respectively. These features are concatenated as $\condition = \conditioncategorical \oplus \conditioncontinuous$, where $\oplus$ represents the vector concatenation operation. Thus, the metadata domain is defined as $\conditioncategoricaldomain \times \conditioncontinuousdomain$. Note that the domains of $\conditioncategorical$ and $\conditioncontinuous$ allow time-varying metadata features.

\emph{Example:} Consider generating time series data representing traffic volume on a highway ($\nchannels = 1$) over 96 hours ($\timeserieslength = 96$) using paired metadata. This metadata includes seven time-varying categorical features like holidays (12 unique labels) and weather descriptions (11 unique labels), denoted by $\numcategoricalfeatures = 7$. It also includes four time-varying continuous features like expected temperature and rain forecast, represented by $\numcontinuousfeatures = 4$.

We denote the dataset $\jointdataset = \{(\timeseries_i, \condition_i)\}_{i=1}^n$ consisting of $n$ independent and identically distributed (i.i.d) samples of time series data $\timeseries$ and paired metadata $\condition$, sampled from a joint distribution $\jointdatadistribution$. \textbf{Our objective is to} develop a conditional generation model $\generationfunction$ such that the samples generated by $\generationfunction(c)$ distributionally match $\conditionaldatadistribution$.

\section{Conditional Time Series Generation using \timeweaver}
\label{sec:timeweaver}
Our approach, \timeweaver, is a diffusion-based conditional generation model. We choose \acp{dm} over \acp{gan} as we consider heterogeneous metadata, \emph{i.e.}, the metadata can contain categorical, continuous, or even time-varying features. Previous works show that the conditional variants of \acp{gan} suffer from mode collapse when dealing with continuous conditions \cite{ding2020ccgan}. Additionally, the proposed alternatives have not been tested in the time series domain for heterogeneous metadata. Our \timeweaver \space model consists of two parts - a denoiser backbone that generates data and a preprocessing module that processes the time-varying categorical and continuous metadata features.

\textbf{Metadata Preprocessing:} The preprocessing step involves handling the metadata $\condition = \conditioncategorical \oplus \conditioncontinuous$. Here, $\conditioncategorical \in \conditioncategoricaldomain$ and $\conditioncontinuous \in \conditioncontinuousdomain$ represent time-varying categorical and continuous metadata features, respectively (see \cref{sec:problem_formulation}). To better incorporate these features from different modalities, we process them separately and then combine them with a self-attention layer. 
\begin{itemize}
    \item The categorical tokenizer $\cattoken$ first converts each category in $\conditioncategorical$ into one-hot encoding and then processes with \ac{fc} layers to create the categorical embedding $\categoricalembeddings \in \cattokendomain$. Similarly, the continuous tokenizer $\conttoken$ also uses \ac{fc} layers to encode continuous metadata $\conditioncontinuous$ into the continuous embedding $\continuousembeddings \in \conttokendomain$. These \ac{fc} layers learn the correlation between metadata features within the categorical and continuous domains. Using \ac{fc} layers is just a design choice; more sophisticated layers can also be used.
    \item $\categoricalembeddings$ and $\continuousembeddings$  are then concatenated and passed through a self-attention layer $\salayer$ to generate the metadata embedding $\conditionalembeddings \in \zdomain $. The self-attention layer equips the generative model to capture the temporal relationship between different metadata features.
\end{itemize}
Here, $d_{\textnormal{cat}}$, $d_{\textnormal{cont}}$, and $d_{\textnormal{meta}}$ are design choices, and we refer the reader to \cref{appendix:time_weaver_arch} for further details.

\textbf{Denoiser:} As the denoiser backbone for \timeweaver, we rely on two state-of-the-art architectures -  CSDI \cite{tashiro2021csdi} and SSSD \cite{alcaraz2022diffusionbased}. The CSDI model uses feature and temporal self-attention layers to process sequential time series data, while SSSD uses structured state-space layers. Note that these denoisers are designed for imputation and forecasting tasks. So, they are designed to take unimputed and historical time series as respective inputs. We modify these denoisers into more flexible metadata-conditioned time series generators by augmenting them with preprocessing layers ($\cattoken$, $\conttoken$, and $\salayer$). We refer the reader to \cref{appendix:time_weaver_arch} for details regarding architectural changes. We train the preprocessing layers and the denoiser $\denoiser$ jointly with the following loss:
\begin{align}
  \loss_{(\denoiser, \conditionembedder, \conttoken, \cattoken)} = & \mathbb{E}_{\textcolor{red}{x,c \sim \jointdataset}, \epsilon \sim \mathcal{N}(\textbf{0}, \textbf{I)},t \sim \mathcal{U}(1,T)} \notag \\
     & \left[ \|  \epsilon - \denoiser(x_t,t,\textcolor{red}{z}) \right \|^2_2], 
      \label{eq:objective}
\end{align}
where $z = \salayer( \cattoken(c_{\textnormal{cat}}) \oplus \conttoken(c_{\textnormal{cont}}))$, $\jointdataset$ represents the dataset of time series and paired metadata sampled from the joint distribution $\jointdatadistribution$, and $T$ is the total number of diffusion steps. As explained in \cref{sec:related_work}, minimizing the loss in \cref{eq:objective} allows \timeweaver \space to learn how to generate samples from the conditional distribution $\conditionaldatadistribution$. During inference, we start from $x_T \sim \mathcal{N}(\mathbf{0}, \mathbf{I})$ and iteratively denoise (with metadata $c$ as input) for $T$ steps to generate $x_0 \sim \conditionaldatadistribution$. This process is depicted in \cref{fig:arch}.

\section{Joint Frechet Time Series Distance}
\label{sec:jftsd}
A good distance metric should penalize the conditional generation approach (provide higher values) if the real and generated joint distributions of the time series and the paired metadata do not match. Existing metrics such as Context-FID \cite{jeha2021psa} rely only on the time series feature extractor, and the metric computation does not involve the paired metadata. This prevents these metrics from penalizing conditional generation approaches for their inability to reproduce metadata-specific features in the generated time series. Therefore, we propose a new metric to evaluate metadata-conditioned time series generation, the \acf{jftsd}. 

In \ac{jftsd}, we compute the \ac{fd} between the real and generated joint distributions of time series and paired metadata. Consider samples from a real data distribution indicated by $\dataset_r = \left\{ (\timeseries_1^r, \condition_1), \ldots, (\timeseries_n^r, \condition_n) \right\}$, where $x_i^r \in \timeseriesdomain$ indicates the time series, and $c_i \in \conditiondomain$ indicates the paired metadata as explained in \cref{sec:problem_formulation}. We denote the dataset of generated time series and the corresponding metadata as $\dataset_g = \left\{ (\timeseries_1^g, \condition_1), \ldots, (\timeseries_n^g, \condition_n) \right\}$, where $\timeseries_i^g = \generationfunction(\condition_i) \ \forall \ i \in [1,n]$, and $\generationfunction$ denotes any arbitrary conditional generation model, as defined in \cref{sec:problem_formulation}. First, similar to the \ac{fid} and \ac{fjd} computations, we project the time series and the paired metadata into a lower-dimensional embedding space using $\metricembedderts(\cdot): \timeseriesdomain \rightarrow \mathbb{R}^{d_{\textnormal{emb}}}$ and $\metricembeddercondn(\cdot):  \conditiondomain \rightarrow \mathbb{R}^{d_{\textnormal{emb}}}$ as the respective feature extractors, where $d_{\textnormal{emb}}$ is the size of the embedding. We concatenate these time series and metadata embeddings to create a joint embedding space. We then calculate the \ac{fd} over the joint embedding space. As such, the \ac{jftsd} metric is formally defined as:
\begin{align} \label{eq:jftsd}
    \begin{split}
    \textnormal{J-FTSD}(\dataset_g, \dataset_r) &= 
    \| \mu_{z^r} - \mu_{z^g} \|^2 \\
    & + \textnormal{Tr}(\Sigma_{z^r} + \Sigma_{z^g} - 2(\Sigma_{z^r} \Sigma_{z^g})^\frac{1}{2}).
    \end{split}
\end{align}
Here, $\mu_{z^d}$ and $\Sigma_{z^d}$ for $d \in \{g,r\}$ are calculated as:
{\small
\begin{align*}
    z^d_i &= \metricembedderts(\timeseries^d_i) \oplus \metricembeddercondn(\condition_i) \quad \quad \forall i : (x_i^d,c_i) \in \dataset_d,\\
     \mu_{z^d} &= \frac{1}{n} \sum_{i=1}^{n} z^d_i, \; \Sigma_{z^d} = \frac{1}{n-1} \sum_{i=1}^{n} 
    (z^d_i - \mu_{z^d})( z^d_i - \mu_{z^d})^\top.
\end{align*}
}
In essence, \ac{jftsd} computes the FD between the Gaussian approximations of the real and generated joint embedding datasets. In \cref{eq:jftsd}, $\mu_{z^r}$, $\mu_{z^g}$ and $\Sigma_{z^r}$, $\Sigma_{z^g}$ are the mean and the variance of the Gaussian approximations of the real and generated joint embedding datasets, respectively. 

\begin{table*}[ht]
\centering
\scalebox{0.95}{
\begin{small}
\begin{sc}
\resizebox{\textwidth}{!}{%
\begin{tabular}{lllll}
\hline
Dataset     & Horizon & \# Channels & Categorical features                                                                                                                                           & Continuous features                                                                                              \\ \hline
Air Quality \cite{misc_beijing_multi-site_air-quality_data_501} & 96      & 6           & \begin{tabular}[c]{@{}l@{}}12 Stations, 5 years, 12 months, \\ 31 dates, 24 hours,  17 wind directions\end{tabular}                                            & \begin{tabular}[c]{@{}l@{}}Temperature, Pressure,\\ Dew point temperature, \\ Rain levels, Wind speed\end{tabular} \\ \hline
Traffic  \cite{misc_metro_interstate_traffic_volume_492}   & 96      & 1           & \begin{tabular}[c]{@{}l@{}}12 holidays, 7 years, 12 months, \\ 31 dates, 24 hours, \\ 11 broad weather descriptions,\\  38 fine weather descriptions\end{tabular} & \begin{tabular}[c]{@{}l@{}}Temperature, Rain levels,\\ Snow fall levels, \\ Cloud conditions\end{tabular}          \\ \hline
Electricity \cite{misc_electricityloaddiagrams20112014_321} & 96      & 1           & 370 users, 4 years, 12 months, 31 dates                                                                                                                        & N.A.                                                                                                               \\ \hline
ECG \cite{wagner2020ptb}        & 1000     & 8          & 71 heart disease statements                                                                                                                                    & N.A.                                                                                                               \\ \hline
\end{tabular}%
}
\end{sc}
\end{small}
}
\vspace{-0.5em}
\caption{\textbf{Dataset overview for experiments with \timeweaver.} This table outlines the key characteristics of the datasets employed in our experiments. These datasets, encompassing Air Quality, Traffic, Electricity, and \ac{ecg}, have been selected to demonstrate \timeweaver's versatility across different time horizons (\textit{col 1}), number of channels (\textit{col 2}), and a wide range of metadata types (\textit{col 3,4}).  }
\label{tab:dataset}
\vspace{-0.5em}
\end{table*}

\textbf{Training Feature Extractors:} Now, we describe our approach to obtain the feature extractors $\metricembedderts$ and $\metricembeddercondn$. As explained in \cref{sec:related_work}, \citet{devries2019evaluation} suggest using separate encoders for data samples and conditions. However, they only deal with a specific type of condition, and this naturally poses a problem for a straightforward extension of their approach to our case, where the metadata could be any arbitrary combination of categorical, continuous, and time-varying features. As such, we propose a novel approach to train the feature extractors $\metricembeddercondn$ and $\metricembedderts$ specific to the time series domain. We jointly train $\metricembedderts$ and $\metricembeddercondn$ with contrastive learning to better capture the joint distribution of the time series and paired metadata as contrastive learning is a commonly used method to map data from various modalities into a shared latent space \cite{Yuan2021MultimodalCT, zhang22acontrastive,ramesh2022hierarchical}.

\begin{figure*}[!ht]
\centering
 \includegraphics[width=1\textwidth, keepaspectratio]{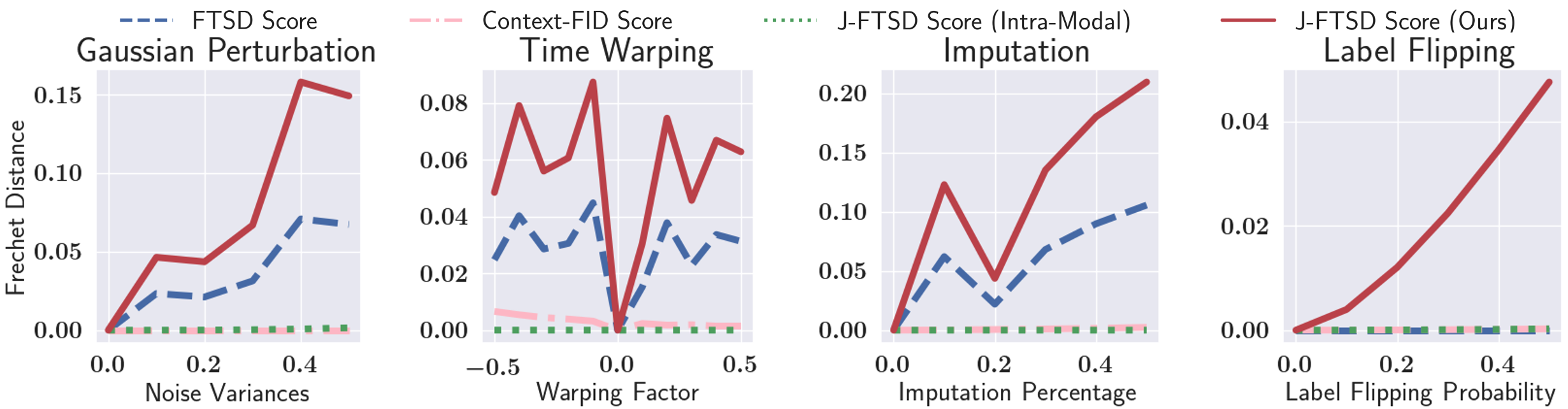}
 \vspace{-2em}
\caption{\textbf{\ac{jftsd} correctly penalizes the conditional time series data distribution.} A good metric should penalize the conditional generation approaches for not being specific to the metadata and deviating from real time series data distribution. As such, we compare the sensitivity of different distance metrics under various synthetic disturbances on the Air Quality dataset (starting from the left); we add Gaussian noise, warp, impute, and randomly change the metadata of the time series samples. We clearly show that as the amount of perturbation increases, our \ac{jftsd} metric (in red) shows the highest sensitivity, correctly capturing the dissimilarities between the perturbed and the original datasets. In contrast, the other metrics remain unchanged or show lower sensitivity.
}
\vspace{-1em}
\label{fig:metric_comparison}
\end{figure*}

\begin{small}
\begin{algorithm}
    \caption{One iteration for training time series $\metricembedderts$ and metadata $\metricembeddercondn$ feature extractors.}
    \begin{algorithmic}[1]
        \STATE {\bfseries Input:}  Time series feature extractor $\metricembedderts$, Metadata feature extractor $\metricembeddercondn$, Time series batch $X_{\textnormal{batch}}$, Paired Metadata batch $C_{\textnormal{batch}}$, Number of patches $\patchsize$, Patch length $\patchlength$, Batch size $\batchsize$. 
        \STATE Randomly select $\patchsize$ patches of length $\patchlength$ from each sample in $X_{\textnormal{batch}}$ and $C_{\textnormal{batch}}$ to generate $X_{\textnormal{batch}}^{\textnormal{patch}}$ and $C_{\textnormal{batch}}^{\textnormal{patch}}$.
        \STATE Obtain the time series and metadata embedding - $\metricembedderts(X_{\textnormal{batch}}^{\textnormal{patch}})$ and $\metricembeddercondn(C_{\textnormal{batch}}^{\textnormal{patch}})$ respectively. 
        
        \STATE Obtain the \texttt{logits} - $\metricembedderts(X_{\textnormal{batch}}^{\textnormal{patch}})^T\metricembeddercondn(C_{\textnormal{batch}}^{\textnormal{patch}})$. 
        
        
        \STATE Define the \texttt{labels} - $[0,1,2,\dots,\batchsize \times \patchsize-1]$.
        \STATE Compute $\mathcal{L}_{\textnormal{time}} = \mathcal{L}_{\textnormal{CE}}(\texttt{logits}, \texttt{labels})$. 
        
        
        \STATE Compute $\mathcal{L}_{\textnormal{meta}} = \mathcal{L}_{\textnormal{CE}}(\texttt{logits.T}, \texttt{labels})$. 
        
        
        \STATE Compute $\mathcal{L}_{\textnormal{total}} = (\mathcal{L}_{\textnormal{time}} + \mathcal{L}_{\textnormal{meta})} / 2$. 
        
        
        \STATE Update parameters of $\metricembedderts$ and $\metricembeddercondn$ to minimize $\mathcal{L}_{\textnormal{total}}$. 
    
    \end{algorithmic}  
\label{alg:metric_training}
\end{algorithm}
\end{small}

\Cref{alg:metric_training} summarizes one training iteration of our feature extractors $\metricembedderts$ and $\metricembeddercondn$. This is visually depicted in \cref{fig:cliplike}. Given the batch of time series $X_{\textnormal{batch}}$ and metadata $C_{\textnormal{batch}}$, we randomly pick $\patchsize$ patches with horizon $\patchlength$ from each time series and metadata sample in batches $X_{\textnormal{batch}}$ and $C_{\textnormal{batch}}$ (line 1). Then, we obtain the time series and metadata embeddings for all patches through their respective feature extractors, $\metricembedderts$ for time series, and $\metricembeddercondn$ for metadata (line 2). Finally, we compute the dot product of time series and metadata embeddings (line 3) and obtain the symmetric cross-entropy loss (lines 4 - 7), which is used to jointly update the parameters of $\metricembedderts$ and $\metricembeddercondn$ (line 8). 

In essence, we learn a joint embedding space for time series and metadata by jointly training $\metricembedderts$ and $\metricembeddercondn$. This is achieved by adjusting the feature extractors' parameters to maximize the cosine similarity of the time series embeddings and the metadata embeddings of $\batchsize \times \patchsize$ pairs of time series and paired metadata in the batch. In our experiments, we used the Informer encoder architecture \cite{zhou2021informer} for $\metricembedderts$ and $\metricembeddercondn$. We choose $\patchlength$ based on the length of the smallest chunk of the time series that contains metadata-specific features. We refer the readers to \cref{appendix:metric_arch} for further details on the choices of $\patchsize$, $\patchlength$, and the encoder architecture.

\paragraph{Why is \ac{jftsd} a good metric for evaluating conditional generation models?} One aspect of the  \ac{jftsd} computation involves estimating the covariance between the time series and the metadata embeddings. Additionally, jointly training the feature extractors with contrastive learning aids in effectively capturing the correlation between the time series and the metadata embeddings. Therefore, the covariance term decreases if the generated time series does not contain metadata-specific features. This allows \ac{jftsd} to accurately penalize for the differences between the real and generated joint distributions, which directly translates to penalizing conditional generation approaches for their poor specificity in reproducing metadata-specific features.

\section{Experiments}
\label{sec:experiments}
We evaluated the performance of \timeweaver \space across datasets featuring a diverse mix of seasonalities, discrete and continuous metadata conditions, a wide range of horizons, and multivariate correlated channels. The list of datasets and their metadata features are provided in \cref{tab:dataset}. All models are trained on the train split, while all metrics are reported on the test split, further detailed in \cref{appendix:dataset_description}.

\begin{table*}
\centering
\scalebox{0.80}{
\begin{small}
\begin{sc}
\begin{tabular}{lcccccccc}
\hline
{Approach} & \multicolumn{2}{c}{Air Quality} & \multicolumn{2}{c}{ECG} & \multicolumn{2}{c}{Traffic} & \multicolumn{2}{c}{Electricity} \\ 
                          & J-FTSD \(\downarrow\) & TSTR \(\uparrow\)      & J-FTSD \(\downarrow\)    & TSTR \(\uparrow\)  & J-FTSD \(\downarrow\)     & TSTR \(\uparrow\)    & J-FTSD \(\downarrow\)         & TSTR \(\uparrow\)       \\ \hline
WaveGAN \\
\cite{donahue2019wavegan}  & 14.25$\pm$0.79 & 0.61$\pm$0.01 & 9.55$\pm$0.01 & 0.65$\pm$0.001 & 25.69$\pm$0.01 & 0.54$\pm$0.01 & 7.82$\pm$0.002 & 0.57$\pm$0.007 \\ \hline
Pulse2Pulse \\
\cite{thambawita2021pulse2pulse}  & 22.07$\pm$0.02 & 0.60$\pm$0.002 & 13.49$\pm$0.04 & 0.63$\pm$0.03 & 17.70$\pm$0.002 & 0.52$\pm$0.03 & 2.8$\pm$0.01 & 0.71$\pm$0.004 \\ \hline
\timeweaver-CSDI   & \textbf{2.2$\pm$0.07} & \textbf{0.77$\pm$0.01} & 7.25$\pm$0.09 & 0.83$\pm$0.001 & 0.53$\pm$0.01 & \textbf{0.66$\pm$0.06} & \textbf{0.6$\pm$0.003} & \textbf{0.78$\pm$0.001}  \\ \hline
\timeweaver-SSSD    & 8.61$\pm$0.18 & 0.63$\pm$0.02 & \textbf{5.43$\pm$0.1} & \textbf{0.85$\pm$0.007} & \textbf{0.36$\pm$0.03} & 0.65$\pm$0.07 & 1.19$\pm$0.008 & 0.77$\pm$0.001  \\ \hline

\end{tabular}
\end{sc}
\end{small}
}
\vspace{-0.5em}
\caption{\textbf{\timeweaver \space outperforms \ac{gan}-based approaches on the \ac{jftsd} and \ac{tstr} metrics}. The table shows the performance of all the models (rows) on specified datasets (columns). The \timeweaver \space variants significantly outperform \acp{gan} on the \ac{jftsd} and TSTR metrics. Our experimental findings also confirm that lower \ac{jftsd} scores correspond to higher \ac{auc} (TSTR) scores when tested on the original test dataset, showcasing the utility of our proposed \ac{jftsd} metric in evaluating the quality of the generated data distribution. We report the mean and standard deviation of both metrics, averaged over three seeds.}
\vspace{-1em}
\label{tab:quant_comp}
\end{table*} 
\textbf{Baselines:} We represent the results for the CSDI and SSSD backbones for \timeweaver \space as \timeweaver-CSDI and \timeweaver-SSSD, respectively. Since there are no existing approaches for metadata-conditioned time series generation with categorical, continuous, and time-variant metadata features, we modify the existing state-of-the-art \ac{gan} approaches to incorporate metadata conditions, similar to \timeweaver. The \ac{gan} baselines include \ac{cnn}-based approaches like WaveGAN \cite{donahue2019wavegan}, an audio-focused \ac{gan} model, and Pulse2Pulse \cite{thambawita2021pulse2pulse}, a model specializing in DeepFake generation. Additionally, we compare the \timeweaver \space variants against a diffusion model with a U-Net 1D \cite{ronneberger2015u} denoiser backbone. We chose the 1D variant of U-Net as the basic U-Net architecture is the most preferred denoiser backbone used in pixel-space image diffusion. The exact training details are provided in \cref{appendix:time_weaver_arch,appendix:gan_baselines}. We additionally tried comparing with TimeGAN \cite{yoon2019TimeGAN}, a Recurrent Neural Network (RNN) based approach, and TTS-GAN \cite{Li2022TTSGANAT}, a Transformer-based approach. However, both of these GAN models did not converge on any of the datasets. We show their training results in \cref{appendix:additional_gan}.

\textbf{Evaluation Metrics:} We evaluate our approaches and the \ac{gan} baselines using the \ac{jftsd} metric, as detailed in \cref{sec:jftsd}. To validate the correctness of \ac{jftsd}'s evaluation, we also report the \acf{auc} scores of a classifier trained only using synthetic data. The classifier is trained to predict the metadata given the corresponding synthetic time series. We then test this classifier on the real unseen test dataset. High accuracy indicates that our synthetic data faithfully retains critical features of the paired metadata. We use a standard ResNet-1D \cite{he2016identity} model for the classifier. We denote this metric as \ac{tstr} in \cref{tab:quant_comp}. For each dataset, the categories for which we train the classifier are listed as follows: Electricity - Months (12), Air Quality - Stations (12), Traffic - Weather Descriptions (11), and ECG - Heart Conditions (71). The exact training steps of the classifiers and a detailed description of the metrics are outlined in \cref{appendix:evaluation_metric}.

\textbf{Experimental Results and Analysis:} Our experiments demonstrate that the \timeweaver \space models significantly outperform baseline models in synthesizing time series data across all evaluated benchmarks. Our experiments address the following key questions:

\textbf{Does the \ac{jftsd} metric correctly penalize when the generated time series samples are not specific to the paired metadata?} 
\begin{figure*}[!ht]
\centering
 \includegraphics[width=0.94\textwidth, keepaspectratio]{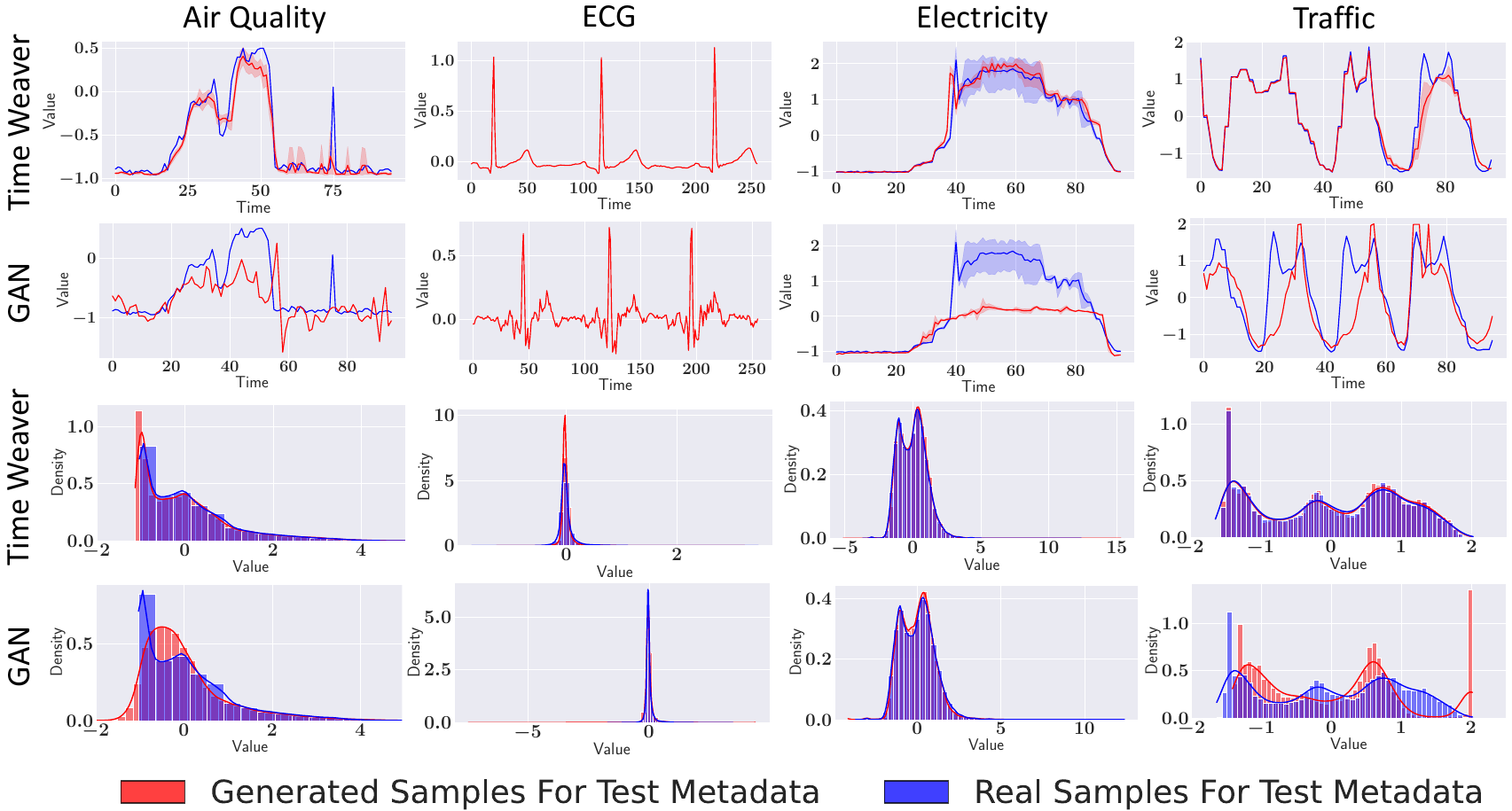}
 \vspace{-1em}
\caption{\textbf{\timeweaver \space generated time series distributions match the real time series distributions.} Each column represents a different dataset. The real time series is in blue, while the generated time series is in red. The first \& third rows correspond to the \timeweaver \space model, and the second \& fourth rows correspond to the best-performing GAN model. The top two rows have the real and generated time series for
unseen test metadata conditions. The bottom two rows compare the histograms of the real and generated time series values aggregated over their respective datasets, also for unseen test metadata conditions. Both results indicate that our \timeweaver \space model can generate realistic time series samples that are specific to the corresponding metadata conditions, beating the previous state-of-the-art GAN model. In both scenarios, the GAN model fails to match the real time series data distribution, while our \timeweaver \space model has learned the correct conditional distribution for specific metadata conditions, specifically for the Air Quality and Traffic datasets.
}
 \vspace{-1em}
\label{fig:qual_synth}
\end{figure*}
In \cref{fig:metric_comparison}, we assess the sensitivity of our \ac{jftsd} metric against previous FD-based metrics. This assessment involves introducing controlled perturbations into the time series to test the sensitivity of the metric. These perturbations include \textit{Gaussian noise} - which introduces Gaussian noise of increasing variance; \textit{time warping}, involving scaling adjustments; \textit{imputation} - imputing the time series with local mean and \textit{label flipping} - where metadata conditions are randomly changed, decoupling them from the time series. An effective metric should demonstrate an increased sensitivity when the real and generated joint distributions of time series and metadata diverge. We compare \ac{jftsd} against three FD-based metrics: 1) the FTSD score, which calculates the FD using only time series embeddings (derived from the $\metricembedderts$ feature extractor); 2) the Context-FID score \cite{jeha2021psa} where $\metricembedderts$ is trained to maximize similarity of similar time series; 3) the \ac{jftsd} (Intra-Modal) score, which differs from \ac{jftsd} in that the time series and metadata feature extractors are trained individually to maximize the embedding similarity for similar samples. Our \ac{jftsd} metric is the most sensitive when compared to other metrics under synthetic disturbances. The key benefit of our metric can be observed in the label-flipping experiment, where only our metric increases as we increase the label-flipping probability in the paired metadata conditions. Other metrics remain unchanged and lack sufficient sensitivity because they overlook paired metadata in their distance calculations, a critical factor that \ac{jftsd} adeptly incorporates. Additionally, the \ac{jftsd} (Intra-Modal) score remains mostly unchanged under these perturbations, highlighting the advantage of the joint training of the time series and metadata feature extractors in our metric. Experiments in \cref{fig:metric_comparison} underscore the importance of our \ac{jftsd} metric in assessing the quality of the generated time series data.

\textbf{How does \timeweaver \space compare against other approaches on real-world datasets?} Across all the datasets, the \timeweaver \space variants consistently outperform \ac{gan} models in terms of \ac{jftsd} scores, as shown in \cref{tab:quant_comp}. Specifically, we beat the best \ac{gan} model by roughly $6\times$ on the Air Quality dataset, $1.75 \times$ on the ECG dataset, $4\times$ on the Electricity dataset, and more than $40\times$ on the Traffic dataset. To understand this performance gain, we conduct an ablation study where we compare the \timeweaver-CSDI model and WaveGAN \cite{donahue2019wavegan} in the presence and absence of metadata on the Air Quality and Electricity datasets. We use the FTSD metric, which is the Frechet Distance without considering the metadata embeddings, as our evaluation metric. The quantitative results of this study are summarized in \cref{tab:metadata_ablation}. We find that \timeweaver \space provides lower FTSD for the Electricity dataset, and WaveGAN provides lower FTSD for the Air Quality dataset in the absence of metadata. However, in the presence of metadata, we note that \timeweaver \space outperforms WaveGAN on both datasets. The key insight is that \timeweaver \space processes metadata more efficiently than \ac{gan}-based approaches to generate metadata-specific time series. This observation aligns with the recent work \cite{ding2020ccgan} that shows \ac{gan}s perform poorly with continuous conditions. The metadata conditions in the time series domain are so complex (a mix of categorical, continuous, and time-varying conditions) that the \ac{gan} models fail to learn a good conditional time series distribution.    

Additionally, we compare \timeweaver \space against a diffusion model baseline with U-Net 1D \cite{ronneberger2015u} as the denoiser backbone. The quantitative results for this comparison are provided in \cref{tab:unet_comparison}. We note that the best-performing \timeweaver \space model outperforms the diffusion model baseline with the U-Net 1D denoiser, with an average improvement of 9\% in TSTR and 17\% in J-FTSD metrics. This superior performance highlights the importance of time series-specific innovations in \timeweaver \space for complex metadata-specific time series generation.

\textbf{Does the synthetic data generated by \timeweaver \space capture metadata-specific features to train an accurate classifier?} 
When training with the generated synthetic time series data, the classifier's accuracy in classifying metadata hinges on the presence of distinct metadata-specific features in the time series. The high TSTR scores in \cref{tab:quant_comp} strongly suggest that the data generated by \timeweaver \space retain the essential characteristics necessary to train classifiers that exhibit high \ac{auc} on real unseen test data. The marked improvement in \ac{tstr} scores with \timeweaver \space when compared to \ac{gan} models demonstrates both the practical value and the superior quality of the synthetic data generated by our model.

\textbf{Does lower \ac{jftsd} correlate with higher \ac{tstr} performance?} The experimental data, as outlined in \cref{tab:quant_comp}, exhibit a clear correlation: lower \ac{jftsd} scores are consistently associated with higher \ac{tstr} scores on the original, unseen test dataset. This correlation is expected as both metrics evaluate the precision of the generated time series relative to the corresponding metadata, as well as the closeness of the real and generated joint distributions. This further underscores the effectiveness of the \ac{jftsd} metric as a reliable indicator to assess the quality of generated data. We further solidify the validity of \ac{jftsd} by showing that lower values of \ac{jftsd} correspond to lower values of the Dynamic Time Warping (DTW) metric, where DTW is computed between the real and generated time series samples. Lower DTW indicates higher similarity between the real and generated time series. 
We refer the reader to \cref{sec:jftsd_dtw_comparison} for further details and experimental results.

\textbf{Does the synthetic data generated by \timeweaver \space qualitatively match the real data?} \Cref{fig:qual_synth} (top two rows) displays the quality and realism of the time series data generated by the best-performing \timeweaver \space model. This figure contrasts generated time series samples with real ones under identical metadata conditions. The comparison demonstrates that the \timeweaver \space model produces time series samples that are highly similar to real samples, effectively mapping metadata to the corresponding time series. In contrast, \ac{gan} baseline models face challenges in generating realistic time series and accurately mapping metadata. A notable example is their performance with \ac{ecg} signals (2nd column): \ac{gan} models only learn to generate a noisy version of the ECG samples while our \timeweaver \space model generates a pristine, realistic sample. We provide additional qualitative samples in \cref{appendix:additonal_qualitative}.

\textbf{Does the synthetic data generated by \timeweaver \space match the real data in terms of density and spread of time series values?}  In \cref{fig:qual_synth} (bottom two rows), we extend our analysis to compare real and generated data distributions across all datasets. This is achieved by transforming the real and generated time series datasets into histograms of their respective values. Take, for instance, the traffic dataset: we aggregate all samples to form a histogram over the raw time series values for both real and generated datasets. The \timeweaver \space model provides a significantly more accurate representation of the real time series distribution than the best-performing \ac{gan} model. \ac{gan} models consistently fail to learn the complex underlying distributions of real data, particularly evident in the Air Quality and Traffic datasets.

\textbf{Does \timeweaver \space retain the causal relationship between the input metadata and the corresponding generated time series?} We analyze \timeweaver's ability to retain causal relationships using the Air Quality dataset. We note that there exist physical models that indicate the effect on the
particulate matter level (one of the time series channels) whenever there is rainfall (metadata input). We show through qualitative examples in \cref{fig:causality} that the generated particulate matter levels adhere to this effect. However, we note that this is not a rigorous proof of causality. We aim to build upon this work to develop time series generative models that can maintain causal relationships between the input metadata and the generated time series.

\textbf{Training and Inference Complexity:} Here, we discuss the training and inference complexity of our approach. For the inference complexity, we make the following observations:
\begin{itemize}
    \item The \timeweaver-CSDI denoiser’s inference time complexity reflects that of any transformer architecture, scaling quadratically with the time series length $L$ and linearly with respect to the hidden dimension size $H$ and the number of diffusion steps $T$. 
    \item The \timeweaver-SSSD denoiser’s inference time complexity reflects that of the S4 model \cite{alcaraz2022diffusionbased}, scaling quadratically with the hidden dimension size $H$ and the number of diffusion steps $T$.
    \item  Our metadata encoder’s inference time complexity is the same as that of the transformer architecture, $\mathcal{O}(L^2K + K^2L)$, where $K$ is the number of metadata conditions. As the metadata encoder only runs once for a single sample generation, its time complexity is not affected by the number of diffusion steps.
\end{itemize}

Additionally, we tabulate the exact time taken to generate a sample for all datasets in \cref{tab:inference_latency}. For training complexity, we show the rate of change of \ac{jftsd} for \timeweaver \space and \ac{gan} models. Specifically, we considered the Air Quality dataset, and the results are shown in \cref{fig:jftsd_training}. As expected, we observe that the rate of decrease of \ac{jftsd} is quicker for \timeweaver \space when compared to \ac{gan} models.

\section{Conclusion}
\label{sec:conclusion}
This paper addresses a critical gap in synthetic time series data generation by introducing \timeweaver, a novel diffusion-based generative model. \timeweaver \space leverages heterogeneous paired metadata, encompassing categorical, continuous, and time-variant variables, to significantly improve the quality of generated time series. Moreover, we introduce a new evaluation metric, \ac{jftsd}, to assess conditional time series generation models. This metric offers a refined approach to evaluate the specificity of generated time series relative to paired metadata conditions. Through \timeweaver, we demonstrate state-of-the-art results for generated sample quality and diversity across four diverse real-world datasets.

\textbf{Limitations:} Despite its superior performance in generating realistic time series data, \timeweaver \space encounters challenges typical of \acp{dm}, including slower inference and prolonged training durations compared to GAN-based models. Future work will focus on overcoming these limitations, potentially through techniques such as progressive distillation \cite{salimans2022progressive} for accelerated inference. We also aim to explore the application of heterogeneous paired metadata conditions to enhance forecasting and anomaly detection within the time series domain.


\newpage


\section*{Impact Statement}
This paper presents work whose goal is to advance the field of Machine Learning. There are many potential societal consequences 
of our work, none of which we feel must be specifically highlighted here.

\bibliography{references/external, references/swarm}
\bibliographystyle{icml2024}

\newpage
\appendix
\onecolumn

\section{Appendix}


\subsection{Diffusion Process} \label{appendix:diffusion_primer}
Diffusion Models (DMs) are trained to denoise a noisy sample, referred to as the reverse or the backward process $p_\theta$. The noisy samples are generated by a Markovian forward process $q$, where we gradually corrupt a clean sample from the dataset by adding noise for $T$ diffusion steps. The forward process is predetermined by specifying a noise schedule $\left\{ \beta_1, \ldots, \beta_T \right \}$, where $\beta_t \in [0,1]$. The following equations parameterize the forward process:
\begin{align}
    &q(x_1, \ldots, x_T \mid x_0)  = \prod_{t=1}^{T} q(x_t \mid x_{t-1}), \\
    &q(x_t \mid x_{t-1}) = \mathcal{N}(\sqrt{1-\beta_t}x_{t-1}, \beta_t \mathbf{I}),
\end{align}
where $\mathcal{X}$ is the data distribution that we want to learn, $x_0 \sim \mathcal{X}$, $\mathcal{N}(\mu, \Sigma)$ represents a Gaussian distribution with mean $\mu$ and covariance matrix $\Sigma$, and $T$ is the number of diffusion steps. The noise schedule $\left\{ \beta_1, \ldots, \beta_T \right \}$ and $T$ are chosen such that $x_T$ resembles samples from a Gaussian distribution with zero mean and unit variance, \emph{i.e.}, $q(x_T) \simeq \mathcal{N}(\mathbf{0}, \mathbf{I})$. This allows us to start the backward process from $x_T \sim \mathcal{N}(\mathbf{0}, \mathbf{I})$ and iteratively denoise for $T$ steps to obtain a sample from $\mathcal{X}$. The reverse process is parameterized as follows:
\begin{align}
    &p_\theta(x_0, \ldots, x_{T-1} \mid x_T)  = p(x_T) \prod_{t=1}^{T} p_\theta(x_{t-1} \mid x_t).
\end{align}
Here, $p(x_T) = \mathcal{N}(\mathbf{0}, \mathbf{I}).$ Essentially, the reverse process is learnable, and $p_\theta(x_{t-1} \mid x_t)$ approximates $q(x_{t-1} \mid x_t, x_0)$. $p_\theta(x_{t-1} \mid x_t)$ is parameterized using a neural network, $\denoiser$. \citet{ho2020denoising} show that through simple reparametrization tricks, we can convert the learning objective from approximating $q(x_{t-1} \mid x_t, x_0)$ to estimating the amount of noise added to go from $x_{t-1}$ to $x_t$. Thus, the diffusion objective is stated as minimizing the following loss function:
\begin{equation}
     \loss_{\text{DM}} = \mathbb{E}_{x \sim \mathcal{X}, \epsilon \sim \mathcal{N}(\textbf{0}, \textbf{I}),t \sim \mathcal{U}(1,T)} \left[ \| \epsilon - \denoiser(x_t,t) \right \|^2_2], \label{eq:dm_objective_appendix}
\end{equation}
where $t \sim \mathcal{U}(1, T)$ indicates that $t$ is sampled from a uniform distribution between 1 and $T$, $\epsilon$ is the noise added to $x_{t-1}$ to obtain $x_t$, and $\denoiser$ takes the noisy sample $x_t$ and the diffusion step $t$ as input to estimate $\epsilon$. This is equivalent to the score-matching techniques \cite{song2019generative,song2021scorebased}. 
\subsection{Dataset Description} \label{appendix:dataset_description}

In this section, we describe in detail the various datasets used in our experiments, their training, validation, and testing splits, and the normalization procedures.  

\subsubsection{Electricity Dataset}
The Electricity dataset consists of power consumption recordings for 370 users over four years from 2011 to 2015. We frame the following task for this dataset - ``Generate the electricity demand pattern for user 257, for the 3rd of August 2011,'' which is a univariate time series. We consider the following features as the metadata: 370 users, four years, 12 months, and 31 dates (\cref{tab:appendix_dataset}). The power consumption is recorded every 15 minutes, so the time series is 96 time steps long. The total number of samples without any preprocessing is 540200. We remove samples with values of 0 for the entire time series, and the resulting total number of samples is 434781. We establish a data split comprising training, validation, and test sets distributed in an 80-10-10 ratio. To obtain the split, we randomly pick 80\% of the 434781 samples and assign them to the training set. The same is repeated for the validation and the test sets. We avoid using the traditional splits proposed by \citet{Du_2023}, as their split creates certain year metadata features in the test set that do not exist in the training set. For example, no month from 2011 exists in the training set split proposed by \citet{Du_2023}. 

\subsubsection{Traffic Dataset}
For traffic volume synthesis, we use the metro interstate traffic volume dataset. The dataset has hourly traffic volume recordings from 2012 to 2018, along with metadata annotations like holidays, textual weather descriptions, and weather forecasts (\cref{tab:appendix_dataset}). Here, we want to generate time series samples for prompts such as ``Given the following weather forecast, synthesize a traffic volume pattern for New Year’s Day.'', which is a univariate time series. The dataset CSV file has 48204 rows containing the traffic volume. We synthesize the traffic volume for a 96-hour window. So, to create a dataset from the CSV file, we slide a window of length 96 with a stride of 24. This gives a total of 2001 time series samples, which we randomly divide into train, validation, and test sets with an 80-10-10 ratio.

\subsubsection{Air Quality Dataset}
This dataset contains hourly air pollutants data from 12 air quality monitoring stations in Beijing. The meteorological data in each air quality site are paired with the weather data from the nearest weather station (please refer to \cref{tab:appendix_dataset} for more details regarding the metadata conditions). Here, the task is to synthesize a multivariate (6 channels) time series given the weather forecast metadata. The dataset has missing values, which we replace with the mean for both continuous metadata and the time series. For categorical metadata, missing values exist only in the wind direction metadata feature, which we fill using an ``unknown'' label. The dataset is split into train, validation, and test sets based on months. The recordings are available from 2013 to 2017, and we have a total of 576 months, of which we randomly picked 460 as train data, 58 as validation data, and 58 as test data. For each month, we slide a window of length 96 with a stride of 24, and this provides a total of 12166 train time series samples, 1537 validation time series samples, and 1525 test time series samples.

\subsubsection{ECG Dataset}
The PTB-XL ECG dataset is a 12-channel (1000 time steps long) time series dataset with 17651 train, 2203 validation, and 2167 test samples. The dataset has annotated heart disease statements for each ECG time series. Here, the goal is to generate ECG time series samples for a specific heart disease statement, which is our metadata. In this work, we use 8 channels instead of 12, as shown in \cite{alcaraz2023diffusion}.
\begin{table}[t]
\begin{sc}
\resizebox{\textwidth}{!}{%
\begin{tabular}{lllll}
\hline
Dataset     & Horizon & \# Channels & Categorical features                                                                                                                                           & Continuous features                                                                                              \\ \hline
Air Quality & 96      & 6           & \begin{tabular}[c]{@{}l@{}}12 Stations, 5 years, 12 months, \\ 31 dates, 24 hours,  17 wind directions\end{tabular}                                            & \begin{tabular}[c]{@{}l@{}}Temperature, Pressure,\\ Dew point temperature, \\ Rain levels, Wind speed\end{tabular} \\ \hline
Traffic     & 96      & 1           & \begin{tabular}[c]{@{}l@{}}12 holidays, 7 years, 12 months, \\ 31 dates, 24 hours, \\ 11 broad weather descriptions,\\  38 fine weather descriptions\end{tabular} & \begin{tabular}[c]{@{}l@{}}Temperature, Rain levels,\\ Snowfall levels, \\ Cloud conditions\end{tabular}          \\ \hline
Electricity & 96      & 1           & 370 users, 4 years, 12 months, 31 dates                                                                                                                        & N.A.                                                                                                               \\ \hline
ECG         & 1000     & 8          & 71 heart disease statements                                                                                                                                    & N.A.                                                                                                               \\ \hline
\end{tabular}%
}
\end{sc}
\vspace{-1em}
\caption{\small{\textbf{Dataset overview for experiments with \timeweaver.} This table outlines the key characteristics of the datasets employed in our experiments. These datasets, encompassing Air Quality, Traffic, Electricity, and \ac{ecg}, have been carefully selected to demonstrate \timeweaver's versatility across different time horizons, number of channels, and a wide range of metadata types.   }}
\label{tab:appendix_dataset}
\end{table}
\subsection{Architecture Description Of The Feature Extractors Used In \ac{jftsd}} \label{appendix:metric_arch}

\begin{figure}[t]
\centering
\label{fig:cliplike}
\includegraphics[width=0.8\textwidth]{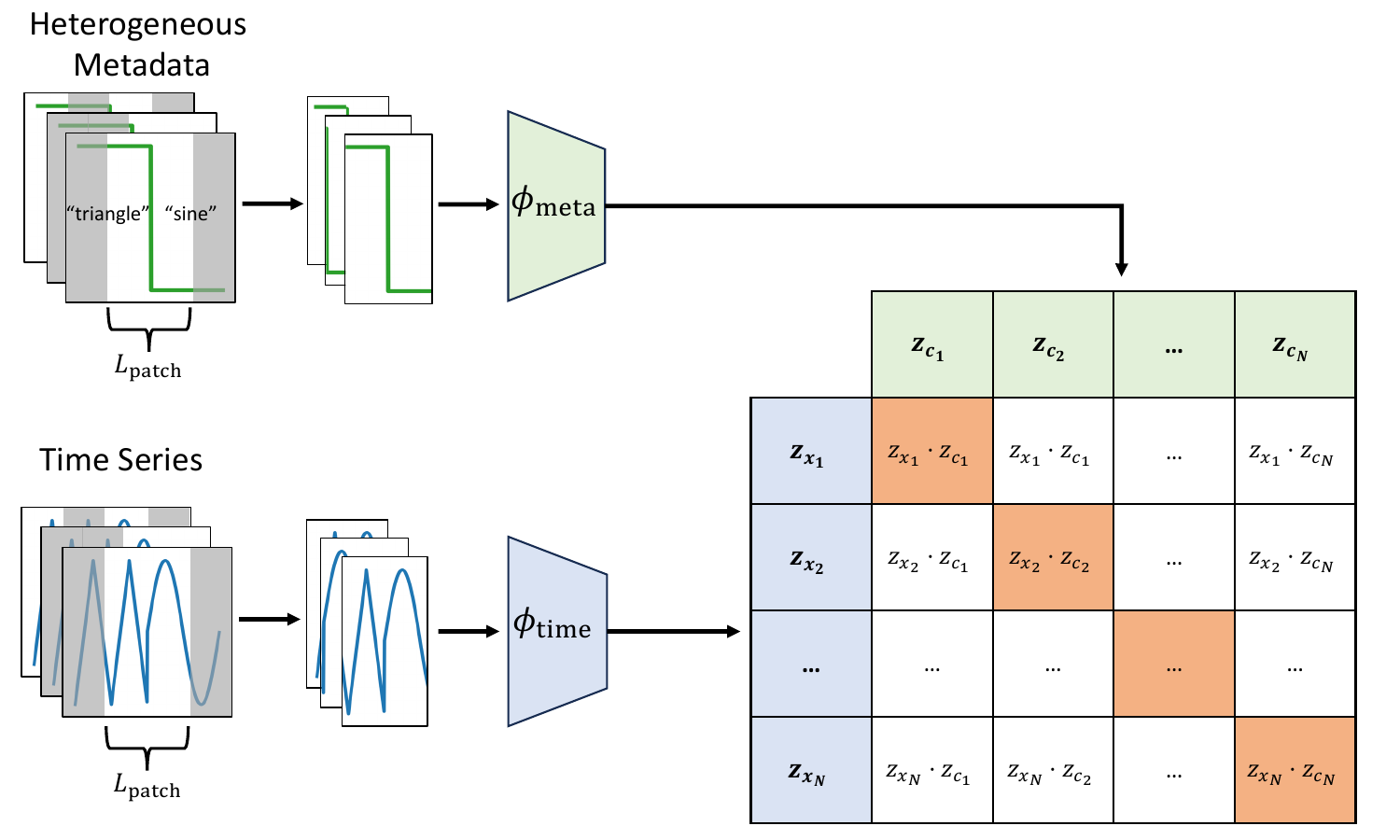}
\vspace{-0.5em}
\caption{\textbf{Contrastive Training of J-FTSD Feature Extractors Inspired By CLIP \cite{radford2021learning}:} This figure depicts the contrastive learning-based training approach for the J-FTSD feature extractors $\metricembedderts$ and $\metricembeddercondn$, akin to the methodology used in CLIP. Here, we consider a time series where the first half is a triangle wave and the second half is a sine wave. The categorical metadata corresponding to this time series has the first half labeled as 1 (``triangle'') and the second half as 0 (``sine''). Patches of length $\patchlength$ are extracted from time series and metadata and processed through their respective feature extractors. The embeddings, $z_{c}$ from the metadata encoder and $z_{x}$ from the time series encoder, are compared using their dot products to identify correct pairings, highlighted along the matrix diagonal (in orange). The feature extractors are trained through contrastive learning, employing cross-entropy loss to enhance the accuracy of matching time series data with its relevant metadata, effectively capturing the nuanced relationship between the two.}
\vspace{-0.01em}
\label{fig:jftsd_training_fig}
\end{figure}

To compute our proposed J-FTSD metric, we relied on the Informer encoder architecture proposed by \citet{zhou2021informer}. Specifically, we used two encoders, one for the time series and one for the metadata features, represented as $\metricembedderts$ and $\metricembeddercondn$, respectively. We made the following modifications to the Informer encoder architecture:

\begin{itemize}
    \item The raw time series was first processed using 1D convolution layers. We added positional encoding to the processed time series before providing it as input to the self-attention layers in $\metricembedderts$. We used the same positional encoding as suggested by \citet{zhou2021informer}.
    \item The raw metadata was processed in the same way as the metadata for the time series generation process, which is highlighted in \cref{sec:timeweaver}. We individually processed or tokenized the categorical and continuous metadata using linear layers and 1D convolution layers to obtain the metadata embedding $\conditionalembeddings$. We added positional encoding to $z$ before providing $z$ as input to the self-attention layers in $\metricembeddercondn$. 
    \item We used 1D convolution layers at the end of every self-attention layer. We used striding after every three self-attention layers, \emph{i.e.}, 1D convolution layers with a stride of 2 were applied after the $3^{rd}$, $6^{th}$, $\ldots$, self-attention layers. 
    \item At the end of the self-attention layers of both $\metricembedderts$ and $\metricembeddercondn$, we flattened the outputs and projected the outputs to a lower-dimensional space using linear layers. We used the Gaussian Error Linear Unit (GELU) activation, similar to the Informer architecture.
\end{itemize}

Now, we describe the choice of $\patchlength$ for each dataset. As explained in \cref{sec:jftsd}, we chose $\patchlength$ based on the minimum horizon required for a patch to contain metadata-specific features. For each dataset, the values of $\patchlength$ and the embedding size (the output dimension of the feature extractors) are tabulated in \cref{tab:jftsd_hyper_dataset}.

\begin{table}[h]
\centering
\begin{sc}
\begin{tabular}{lll}
\hline
\textbf{Dataset}     & \textbf{$\patchlength$} & \textbf{Embedding size} \\ \hline
Air Quality & 64           & 128           \\ \hline
ECG         & 256          & 256            \\ \hline
Electricity & 64           & 48             \\ \hline
Traffic     & 64           & 48             \\ \hline
\end{tabular}
\end{sc}
\caption{\textbf{Choice of the patch and embedding sizes.} In this table, we list the choices of the patch size $\patchlength$ and the embedding size used for each dataset in our experiments.}
\label{tab:jftsd_hyper_dataset}
\end{table}

Specifically, we chose the embedding size such that given a time series sample $x \in \timeseriesdomain$, where $L$ is the horizon, and $F$ is the number of channels in the time series, the embedding size should be smaller than $F \times \patchlength$. This is to ensure that we are reducing the dimensionality of the time series patch. 

Now, we list the hyperparameter choices used for training the feature extractors in \cref{tab:jftsd_hyper_training}. These include the number of patches from a single time series sample $\patchsize$, learning rate, etc, and the design choices in terms of the number of self-attention layers, number of transformer heads, etc. 

\cref{fig:jftsd_training_fig} provides a pictorial representation of the contrastive learning framework used to train the feature extractors $\metricembedderts$ and $\metricembeddercondn$. The training process ensures that the time series samples with similar paired metadata have similar projections in the latent space.

\begin{table}[t]
\centering
\begin{sc}
\begin{tabular}{lll}
\hline
\textbf{Design Parameter} & \textbf{Value} \\ \hline
Embedding size (\textnormal{\texttt{dmodel}}) & 128 \\ \hline
Attention heads (\textnormal{\texttt{nheads}}) & 8 \\ \hline 
Self-attention layers  & 8 \\ \hline 
dropout & 0.05 \\ \hline 
Activation & gelu \\ \hline 
$\patchsize$ & 2 \\ \hline 
Learning Rate & $10^{-4}$ \\ \hline 

\end{tabular}
\end{sc}
\caption{\textbf{Hyperparameters for the feature extractors.}}
\label{tab:jftsd_hyper_training}
\end{table}
\subsection{\timeweaver \space Architecture Design}
\label{appendix:time_weaver_arch}

As mentioned in \cref{sec:timeweaver}, \timeweaver \space has two denoiser backbones - CSDI \cite{tashiro2021csdi} and SSSD \cite{alcaraz2023diffusion}. In this section, we describe the architecture changes we introduced to the CSDI and the SSSD backbones to extend their capabilities to metadata-specific time series generation. 

\subsubsection{\timeweaver-CSDI}
\begin{figure}[H]
\centering
\includegraphics[width=0.8\textwidth]{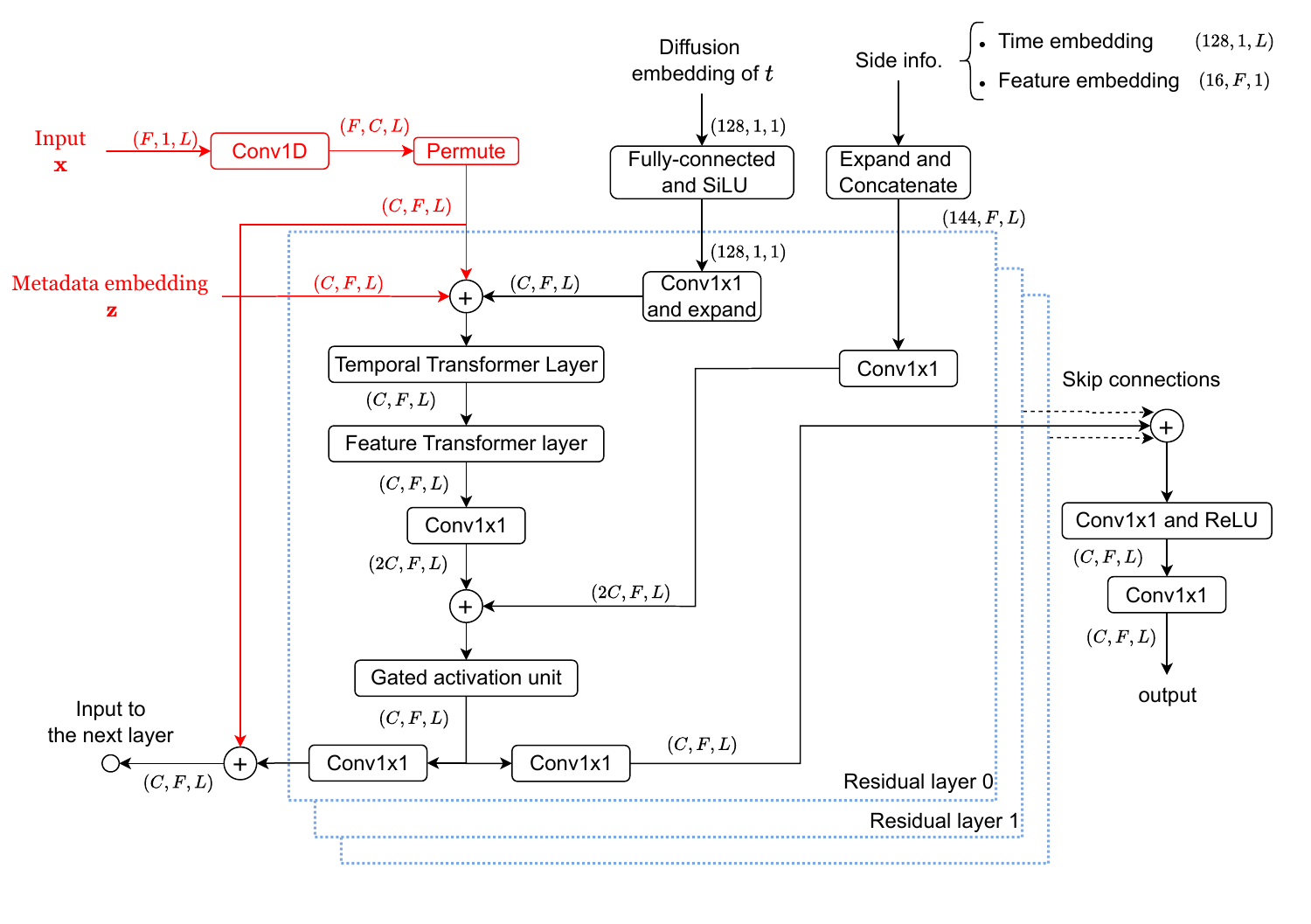}
\vspace{-0.5em}
\caption{\textbf{\timeweaver-CSDI architecture:} This figure shows our changes to the original conditional CSDI model \cite{tashiro2021csdi}. We use this model as our denoiser ($\denoiser$) in our architecture, with metadata preprocessing fixed as in Fig. \ref{fig:arch}. Changes to the original architecture are colored in red. }
\vspace{-1em}
\label{fig:csdi}
\end{figure}

\label{app:csdi}

Consider a batch of time series samples of size $(\batchsize, \nchannels, \timeserieslength)$, where $\batchsize$ represents the number of samples per batch, $\nchannels$ represents the number of channels in the time series, and $\timeserieslength$ represents the horizon. The paired metadata is represented as $ \conditioncategorical \oplus \conditioncontinuous $, where the shape of $\conditioncategorical$ is $(\batchsize, \timeserieslength, \numcategoricalfeatures)$ and the shape of $\conditioncontinuous$ is $(\batchsize, \timeserieslength, \numcontinuousfeatures)$.

\begin{itemize}
    \item \emph{Input time series projection:} We first transformed the input time series batch to $(\batchsize \times \nchannels, 1, \timeserieslength)$ and applied 1D convolution layers with $d_{\textnormal{meta}}$ filters to obtain a projection of shape $(\batchsize \times \nchannels, d_{\textnormal{meta}}, L)$. We then reshaped the projection from $(\batchsize \times F, d_{\textnormal{meta}}, L)$ to $(\batchsize, d_{\textnormal{meta}}, F, L)$.
    \item \emph{Metadata projection:} Simultaneously, we converted each categorical metadata feature in $\conditioncategorical$ into one-hot encoding and further processed using $\cattoken$. Similarly, we processed the continuous metadata, $\conditioncontinuous$, using $\conttoken$. Both $\conditioncategorical$ and $\conditioncontinuous$ were projected to latent representations of shape $(\batchsize, L, d_{\textnormal{cat}})$ and $(\batchsize, L, d_{\textnormal{cont}})$, respectively. These latent representations were concatenated along the final axis and processed using the self-attention layer $\salayer$. At the end of this preprocessing, the categorical and continuous metadata were projected to a latent representation of shape $(\batchsize, L, d_{\textnormal{meta}})$. We then broadcasted and reshaped the projected metadata to  $(\batchsize, d_{\textnormal{meta}}, F, L)$.
    \item \emph{Diffusion step representation:} The CSDI architecture represents the diffusion step using a 128-dimensional representation, which is projected to $d_{\textnormal{meta}}$. We broadcasted and reshaped the diffusion step representation to $(\batchsize, d_{\textnormal{meta}}, F, L)$.
    \item Then, we added the input time series projection, metadata projection, and diffusion step representation. This sum was passed to the temporal and feature transformer layers in the first residual layer. 
    \item We provided the projected metadata as input to all the residual layers in the same manner.
\end{itemize}

These modifications are highlighted in red in \cref{fig:csdi}. For the diffusion process, our experiments with \timeweaver-CSDI used 200 diffusion steps with the noise variance schedule values of $\beta_1 = 0.0001$ and $\beta_T = 0.1$.

Now, we list the architectural choices and the corresponding hyperparameter choices in \cref{tab:csdi_hyperparameters}. The number of residual layers used varies for each dataset. For the Air Quality dataset, we used 10 residual layers. Similarly, for the Traffic, Electricity, and ECG datasets, we used 8, 6, and 10 residual layers, respectively.

\begin{table}[t]
\centering
\begin{sc}
\begin{tabular}{lll}
\hline
\textbf{Design Parameter} & \textbf{Value} \\ \hline
Position embedding & 128 \\ \hline 
Feature or Channel embedding & 16 \\ \hline
Diffusion step embedding & 256 \\ \hline
Embedding size ($d_{\textnormal{meta}}$) & 256 \\ \hline
Attention heads (\textnormal{\texttt{nheads}}) & 16 \\ \hline 
Metadata encoder ($\salayer$) embedding size & 256 \\ \hline 
Metadata encoder ($\salayer$) attention heads & 8 \\ \hline 
Metadata encoder ($\salayer$) Self-attention layers & 2 \\ \hline
Learning rate & $10^{-4}$ \\ \hline
\end{tabular}
\end{sc}
\caption{\textbf{Hyperparameters for \timeweaver-CSDI architecture.}}
\label{tab:csdi_hyperparameters}
\end{table}

\subsubsection{\timeweaver-SSSD} 
\begin{figure}[ht]
\centering
\includegraphics[width=0.8\textwidth]{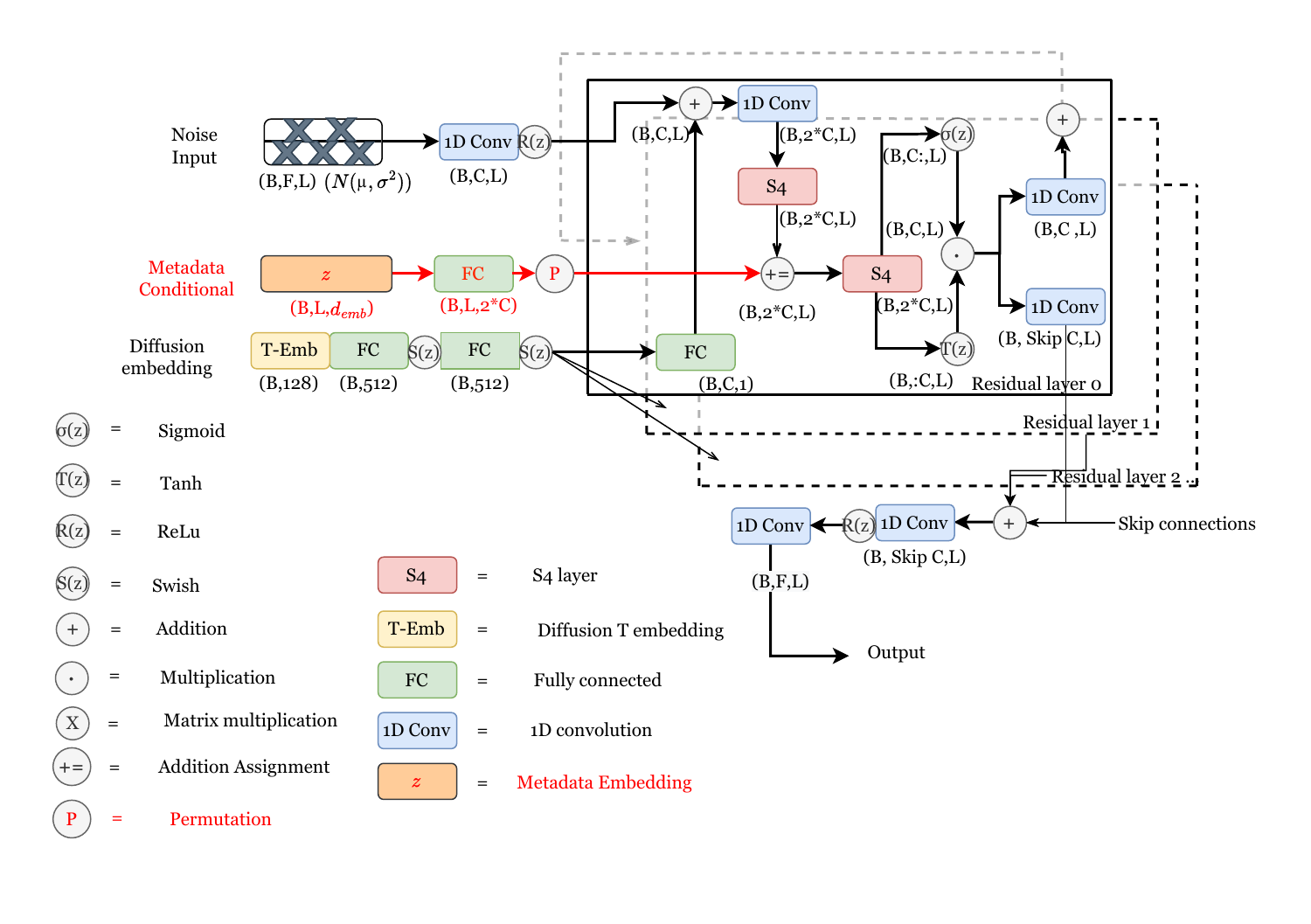}
\vspace{-0.5em}
\caption{\textbf{\timeweaver-SSSD architecture:} This figure shows our changes to the original conditional SSSD model \cite{alcaraz2023diffusion}. We use this model as the denoiser ($\denoiser$) in our architecture, with metadata preprocessing being fixed as in Fig. \ref{fig:arch}. Changes to the original architecture are highlighted in red.  }
\vspace{-1em}
\label{fig:sssd}
\end{figure}
\label{app:sssd}
The \timeweaver-SSSD model is based on the structured state-space diffusion (SSSD)  model \cite{alcaraz2022diffusionbased} which was originally designed for imputation tasks. The SSSD model is built on the DiffWave \cite{kong2021diffwave}  architecture. Unlike the DiffWave model, SSSD utilizes the structured state-space model (SSM) \cite{gu2022efficiently}, which connects the input sequence $u(t)$ to the output sequence $y(t)$ via the hidden state $x(t)$. This relation can be explicitly given as:
\begin{equation*}
    x'(t) = A x(t) + B u(t) \quad \text{and} \quad  y(t) = C x(t) + D u(t). 
\end{equation*}
Here, $A, B, C,$ and $D$ are learnable transition matrices. \citet{gu2022efficiently} propose stacking several SSM blocks together to create a Structured State Space sequence model (S4). These SSM blocks are connected with normalization layers and point-wise \ac{fc} layers in a way that resembles the transformer architecture. This architectural change is done to capture long-term dependencies in time series data. \citet{alcaraz2023diffusion} adjust this architecture to take label input, a binary vector of length 71. As shown in Fig. \ref{fig:sssd}, we replaced this label input with the metadata embeddings obtained from our metadata preprocessing block to incorporate more complex metadata modalities. 

For the diffusion process, our experiments with \timeweaver-SSSD used 200 diffusion steps with the noise variance schedule values of $\beta_1 = 0.0001$ and $\beta_T = 0.02$.

Now, we provide the list of design choices and hyperparameters used in the \timeweaver-SSSD model in \cref{tab:sssd_hyperparameters}. 

\begin{table}[H]
\centering
\begin{sc}
\begin{tabular}{lll}
\hline
\textbf{Design Parameter} & \textbf{Value} \\ \hline
Residual Layer Channels & 256 \\ \hline 
Skip Channels & 16 \\ \hline
Diffusion step embedding input channels & 128 \\ \hline
Diffusion step embedding mid channels & 512 \\ \hline 
Diffusion step embedding output channels & 512 \\ \hline 
S4 layer State dimension & 64 \\ \hline 
S4 layer dropout & 0.0 \\ \hline
Is S4 layer bidirectional & True \\ \hline
Use layer normalization & True \\ \hline
Metadata encoder ($\salayer$) embedding size & 256 \\ \hline 
Metadata encoder ($\salayer$) attention heads & 8 \\ \hline 
Metadata encoder ($\salayer$) Self-attention layers & 2 \\ \hline
Learning rate & $10^{-4}$ \\ \hline
\end{tabular}
\end{sc}
\caption{\textbf{Hyperparameters for the \timeweaver-SSSD architecture.}}
\label{tab:sssd_hyperparameters}
\end{table}

\subsection{\ac{gan} Baselines} \label{appendix:gan_baselines}
\subsubsection{Main \ac{gan} Baselines}
For our main \ac{gan} baselines, we use Pulse2Pulse \ac{gan} \cite{thambawita2021pulse2pulse} and Wave\ac{gan} \cite{donahue2019wavegan}. Since these approaches are not fundamentally conditional, we added additional layers to enable conditional generation. 
\begin{itemize}
    \item For the Electricity and ECG datasets, we used the implementation provided by \cite{thambawita2021pulse2pulse} and \cite{alcaraz2023diffusion}. Since these datasets only have categorical metadata, we represented each categorical label by a fixed embedding. This fixed embedding is added to the output of each layer in the generator after the batch normalization layers. Similarly, we add the fixed embedding to the output of each layer in the discriminator. To learn the conditional distribution, along with predicting whether a sample is real or fake, we also predict the logit of each categorical metadata, similar to \cite{odena2017conditional}. In our training experiments, we noticed that predicting the metadata category for the fake sample results in poor-quality samples. Hence, during training, we only predict the category for the real samples.
    \item For the Air Quality and Traffic datasets, we appended the inputs to the generator and discriminator with the metadata conditions.

\end{itemize}

For all the datasets except the Air Quality dataset, we used min-max normalization to transform the time series samples to lie between -1 and 1. For the Air Quality dataset, we used the standard zero mean, unit variance normalization.

\subsubsection{WaveGAN Implementation Details}
We train the WaveGAN model for all the datasets with a learning rate of $10^{-4}$ and store the checkpoints after every 100 epochs. We sample a $48$-dimensional random vector for the Electricity, Air Quality, and Traffic datasets and a $100$-dimensional random vector for the ECG dataset. This vector serves as the noise input to the generator. We used the \texttt{PyTorch} implementation (\href{https://github.com/mostafaelaraby/wavegan-pytorch/tree/master}{Link to the repo}) and the code from \cite{alcaraz2023diffusion} to implement WaveGAN. We adjusted the number of parameters in the generator and discriminator to roughly match the size of our \timeweaver \space models.
\begin{itemize}
    \item For the Air Quality dataset, the total number of trainable parameters in the GAN model is 15.2 million, and the generator has 8.51 million trainable parameters.
    \item For the Traffic dataset, the total number of trainable parameters in the GAN model is 13.7 million, and the generator has 7.017 million trainable parameters.
    \item For the Electricity dataset, the total number of trainable parameters in the GAN model is 13.3 million, and the generator had 7.17 million parameters.
    \item For the ECG dataset, the total number of trainable parameters in the GAN model is 40.9 million, and the generator has 21.36 million parameters.
\end{itemize}

We chose the checkpoints that provided lower values of \ac{jftsd} on the test data set.

\subsubsection{Pulse2Pulse GAN Implementation Details}
We train the Pulse2Pulse GAN model in the same manner as WaveGAN for all the datasets, with a learning rate of $10^{-4}$, and store the checkpoints after every 100 epochs. Here, the noise input to the generator has the same dimensions as the time series sample that we want to generate. We adjusted the number of parameters in the generator and discriminator to roughly match the size of our \timeweaver \space models. 
\begin{itemize}
    \item For the Air Quality and Traffic datasets, the total number of trainable parameters in the GAN model is 14.1 million, and the generator has 7.45 million trainable parameters.
    \item For the Electricity dataset, the total number of trainable parameters in the GAN model is 16.9 million, and the generator has 8.4 million parameters.
    \item For the ECG dataset, the total number of trainable parameters in the GAN model is 43 million, and the generator has 23.47 million parameters.
\end{itemize}

\subsubsection{Additional \ac{gan} Baselines}
\label{appendix:additional_gan}
\begin{figure}[H]
\centering
 \includegraphics[width=1.0\textwidth]{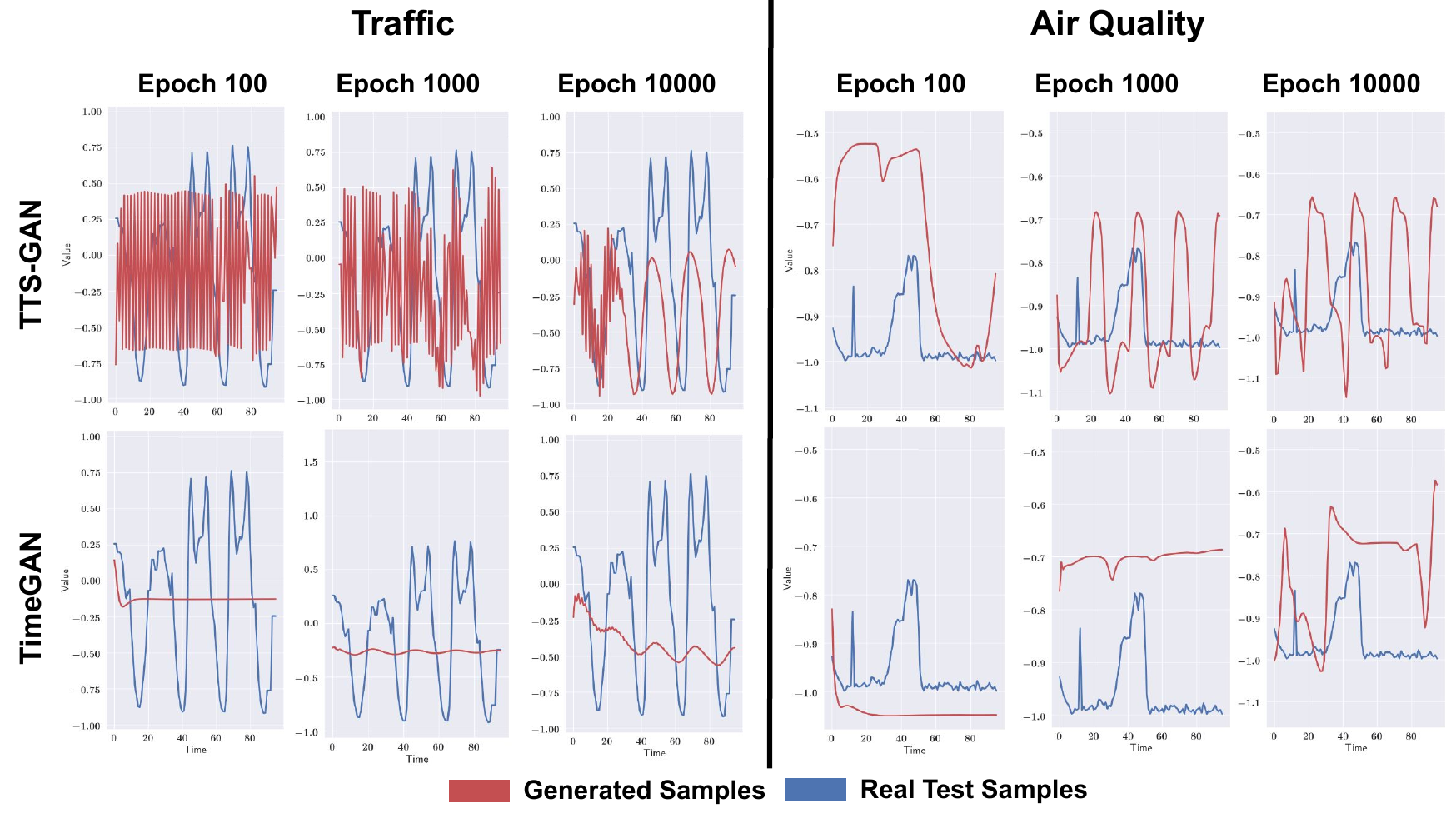}
\caption{\textbf{TimeGAN and TTS-GAN failed to generate realistic samples after 10000 epochs.} This figure shows the samples generated from the test dataset after 100, 1000, and 10000 training epochs, where \textit{row 1} and \textit{row 2} correspond to TTS-GAN and TimeGAN, respectively. We can see that both models fail to generate high-quality realistic samples.
}
\label{fig:timegan}
\end{figure}

In addition to WaveGAN \cite{donahue2019wavegan} and Pulse2Pulse GAN \cite{thambawita2021pulse2pulse} models, we experimented with TTS-GAN \cite{Li2022TTSGANAT} and the well-established TimeGAN \cite{yoon2019TimeGAN} models. Unfortunately, we were unable to train these models to generate quality samples. These models were likely challenged by higher input time series lengths than their original implementation. For example, TimeGAN and TTS-GAN consider time steps up to 24 and 188, respectively, while we consider time steps up to 1000. A similar problem was also faced by \citet{alcaraz2023diffusion}. We include our training examples after 10000 epochs for the Air Quality and Traffic datasets in \cref{fig:timegan}. 

\subsection{Evaluation Metrics} \label{appendix:evaluation_metric}
In this section, we briefly describe the details regarding the evaluation metrics, i.e., TSTR (train on synthetic test on real) and \ac{jftsd}.

\subsubsection{\ac{jftsd} Details}
For the Electricity, Air Quality, and Traffic datasets, the horizon is 96, \emph{i.e.}, $L = 96$. So, we consider time series and metadata patches of length $\patchlength=64$. Consequently, we obtain the time series and the metadata embeddings for these patches using $\metricembedderts$ and $\metricembeddercondn$, respectively. We compute the \ac{jftsd} metric from these embeddings using \cref{eq:jftsd}. For the ECG dataset, since the horizon is 1000, $\patchlength$ is set to 256.

One important detail is that the feature extractors, $\metricembedderts$, and $\metricembeddercondn$, are trained on the entire dataset. This was done to ensure that the feature extractors could effectively obtain the metadata-specific features, improving the evaluation process. For example, if the feature extractors were trained only on the train and validation sets, they would not be able to detect unique trends in the test set.


\subsubsection{Train on Synthetic Test on Real (TSTR) Metric Description}
For TSTR, we use a standard ResNet 1D \cite{he2016identity} architecture. We pick the most physically relevant categorical metadata category for the classification task for each dataset. To this end, we performed the following classification tasks: 
\begin{itemize}
    \item Classification over 12 months in the Electricity dataset. We trained the classifier with cross-entropy loss for 30 epochs with a learning rate of $10^{-4}$.
    \item Classification over 71 heart disease statements in the ECG dataset. We note that for a given time series sample, more than one class could be active. So, we trained a classifier with binary cross-entropy loss for 200 epochs with a learning rate of $10^{-4}$.
    \item Classification over 11 coarse weather descriptions in the Traffic dataset. We treat the classification task here as a multi-class, multi-label classification problem. So, we trained a classifier with binary cross-entropy loss for 250 epochs with a learning rate of $10^{-4}$.
    \item Classification over 12 weather stations for the Air Quality dataset. We used the cross-entropy loss for 500 epochs with a learning rate of $10^{-4}$.
\end{itemize}

Here, we note that with the pre-trained diffusion model, we generated the synthetic train, validation, and test datasets. The classifier is trained on the synthetic train dataset, and the checkpoints are stored with the synthetic validation dataset. We finally evaluated the model on the real test dataset. We use the Area Under the Receiver Operating Characteristic Curve (ROC AUC) as the TSTR metric.
\subsection{Additional Quantitative Results}
This section provides additional quantitative results and ablation studies for \timeweaver \space and \ac{jftsd}.

\subsubsection{Performance Comparison Between \timeweaver \space And \ac{gan} Models In The Presence And Absence Of Metadata}
We perform an ablation study to compare the performance of \timeweaver \space and \ac{gan} in the presence and absence of metadata for the Air Quality and Electricity datasets. We observe the FTSD metric (Frechet Distance without the metadata embedding) for \timeweaver-CSDI and WaveGAN \cite{donahue2019wavegan}. The FTSD metric measures the real and generated time series datasets' distributional similarity. First, we note that in the absence of metadata, WaveGAN provides lower FTSD for the Air Quality dataset, and \timeweaver-CSDI provides lower FTSD for the Electricity dataset. However, in the presence of metadata, \timeweaver-CSDI outperforms WaveGAN in both datasets. The quantitative results are tabulated in \cref{tab:metadata_ablation}. This result aligns with our key insight that \ac{gan}s perform poorly with continuous conditions. However, diffusion-based approaches such as \timeweaver \space can handle any arbitrary combination of categorical, continuous, and time-varying metadata conditions.
\begin{table}[h!]
\centering
\begin{sc}
\begin{tabular}{lllll}
\hline
     Approach       & \multicolumn{2}{l}{Air Quality}                                                                                                          & \multicolumn{2}{l}{Electricity}                                                                                                          \\ \hline
            & \multicolumn{1}{l}{\begin{tabular}[c]{@{}l@{}}Without\\ Metadata\end{tabular}} & \begin{tabular}[c]{@{}l@{}}With\\ Metadata\end{tabular} & \multicolumn{1}{l}{\begin{tabular}[c]{@{}l@{}}Without\\ Metadata\end{tabular}} & \begin{tabular}[c]{@{}l@{}}With\\ Metadata\end{tabular} \\ \hline
WaveGAN \cite{donahue2019wavegan}    & \multicolumn{1}{l}{\textbf{1.40$\pm$0.045}}                                             & 2.62$\pm$0.001                                          & \multicolumn{1}{l}{5.26$\pm$0.025}                                             & 0.92$\pm$0.005                                          \\ \hline
\timeweaver-CSDI & \multicolumn{1}{l}{2.98$\pm$0.097}                                             & \textbf{0.51$\pm$0.016}                                          & \multicolumn{1}{l}{\textbf{0.57$\pm$0.019}}                                             & \textbf{0.29$\pm$0.001}                                          \\ \hline
\end{tabular}
\end{sc}
\caption{\textbf{\timeweaver \space outperforms \ac{gan}s for metadata-specific time series generation for complex metadata.} Here, we show a quantitative comparison between the performance of \timeweaver \space and \ac{gan} models in the presence and absence of metadata. We use the FTSD metric (Frechet Distance without metadata embeddings) for this comparison. We evaluate \timeweaver-CSDI and WaveGAN \cite{donahue2019wavegan} on the Air Quality and Electricity datasets. The key observation is that in the presence of metadata, \timeweaver-CSDI significantly outperforms WaveGAN \cite{donahue2019wavegan} on the FTSD metric. This indicates \timeweaver's ability to handle any arbitrary combination of categorical, continuous, and time-varying metadata.}
\label{tab:metadata_ablation}
\end{table}

\subsubsection{Performance Comparison Between \timeweaver \space And Diffusion Model With U-Net 1D Denoiser}
In addition to \ac{gan}s, we compare the performance of \timeweaver \space against diffusion models with U-Net 1D \cite{ronneberger2015u} as the denoiser backbone. We choose U-Net 1D as U-Net is the most commonly used denoiser in image and video synthesis. This comparison aims to highlight the requirement of time series-specific modifications to the denoiser to obtain high-quality samples. \cref{tab:unet_comparison} shows the quantitative comparison between U-Net 1D and the best-performing \timeweaver \space model for all datasets. We choose the best-performing \timeweaver \space model using the TSTR metric in \cref{tab:quant_comp}. Overall, we observe that \timeweaver \space significantly outperforms the U-Net 1D baseline by around 9\% on the TSTR metric on average. Similarly, \timeweaver \space outperforms the U-Net 1D baseline by around 17\% on the \ac{jftsd} metric.

\begin{table}[]
\centering
\begin{sc}
\begin{tabular}{lllll}
\hline
            &                         & TSTR ($\uparrow$)                  &                        &                         \\ \hline
Approach    & Air Quality             & ECG                    & Traffic                & Electricity             \\ \hline
U-Net 1D \cite{ronneberger2015u}     & 0.66$\pm$0.01           & 0.71$\pm$0.01          & 0.65$\pm$0.02          & \textbf{0.78$\pm$0.003} \\ \hline
\timeweaver & \textbf{0.77$\pm$0.01}  & \textbf{0.85$\pm$0.007} & \textbf{0.66$\pm$0.06} & \textbf{0.78$\pm$0.001} \\ \hline
            &                         & \ac{jftsd} ($\downarrow$)                 &                        &                         \\ \hline
            & Air Quality             & ECG                    & Traffic                & Electricity             \\ \hline
U-Net 1D \cite{ronneberger2015u}     & 7.32$\pm$0.04           & 12.47$\pm$0.07         & \textbf{0.19$\pm$0.01} & 0.64$\pm$0.003          \\ \hline
\timeweaver & \textbf{2.2$\pm$0.07} & \textbf{5.43$\pm$0.1} & 0.53$\pm$0.01          & \textbf{0.6$\pm$0.003} \\ \hline
\end{tabular}
\end{sc}
\caption{\textbf{\timeweaver \space outperforms the diffusion model with a U-Net 1D denoiser backbone.} Here, we show quantitative results for the comparison between the best-performing \timeweaver \space model and the diffusion model baseline with U-Net 1D as the denoiser backbone. Overall, we observe that \timeweaver \space significantly outperforms the U-Net 1D baseline by around 9\% on the TSTR metric and around 17\% on the \ac{jftsd} metric on average. This superior performance highlights the necessity for time series-specific changes to the denoiser backbone in diffusion models. For example, \timeweaver-CSDI contains feature and temporal transformer layers that can learn the required correlation between the input metadata and the generated time series channels, as well as the correlation between different time stamps. Meanwhile, U-Net, a popular denoiser used for image and video generation, cannot effectively capture the temporal and channel-wise correlation between the metadata and time series.}
\label{tab:unet_comparison}
\end{table}

\subsubsection{Correlation Between Our Proposed \ac{jftsd} Metric And The Dynamic Time Warping Metric}
\label{sec:jftsd_dtw_comparison}
Dynamic Time Warping (DTW) \cite{1162641} is a similarity metric between two different time series samples. Dynamic Time Warping looks for the temporal alignment that minimizes the Euclidean distance between two aligned time series samples. Lower values of DTW indicate high similarity between the two time series samples. As a metric to evaluate metadata-specific time series generation models, DTW works effectively when there is only one time series per metadata condition. In this case, we can compare the single real time series with the generated time series for the same metadata to check if the metadata-specific features are retained. If there are multiple time series per metadata condition (which is the case in the ECG dataset), DTW might not be the most effective solution. In our experiments, we note that there is only one time series sample per metadata condition in the Air Quality, Electricity, and Traffic datasets. For these datasets, we observe that, among different approaches used in our experiments, lower values of \ac{jftsd} correspond to lower values of DTW. The quantitative results are provided in \cref{tab:dtw}. This further affirms the validity of our proposed \ac{jftsd} metric.
\begin{table}[]
\centering
\begin{sc}
\begin{tabular}{llll}
\hline
Approach        & Air Quality            & Traffic                & Electricity            \\ \hline
WaveGAN \cite{donahue2019wavegan}        & 4.56$\pm$1.64          & 4.00$\pm$0.66          & 3.79$\pm$4.55          \\ \hline
Pulse2Pulse \cite{thambawita2021pulse2pulse} & 5.45$\pm$2.14          & 4.86$\pm$1.16          & 3.54$\pm$3.63          \\ \hline
\timeweaver-CSDI            & \textbf{2.50$\pm$1.12} & 1.34$\pm$0.99          & \textbf{2.05$\pm$1.64} \\ \hline
\timeweaver-SSSD            & 3.88$\pm$2.20          & \textbf{1.10$\pm$1.00} & 2.61$\pm$2.36          \\ \hline
\end{tabular}
\end{sc}
\caption{\textbf{Our proposed \ac{jftsd} metric correlates with the Dynamic Time Warping metric \cite{1162641}.} This table shows the Dynamic Time Warping (DTW) metric computed between the real and generated time series samples for the Air Quality, Electricity, and Traffic datasets. Note that for the DTW metric computed between real and generated time series samples, lower values indicate higher similarity between the two samples. Comparing with the \ac{jftsd} values in \cref{tab:quant_comp}, we note that lower values of \ac{jftsd} correspond to lower values of DTW. This further affirms the validity of our proposed \ac{jftsd} metric. Additionally, note that our proposed \timeweaver \space models provide the lowest DTW for all the datasets.}
\label{tab:dtw}
\end{table}
\subsection{Training and Inference Complexity}

\begin{figure}[H]
\centering
\includegraphics[width=0.5\columnwidth]{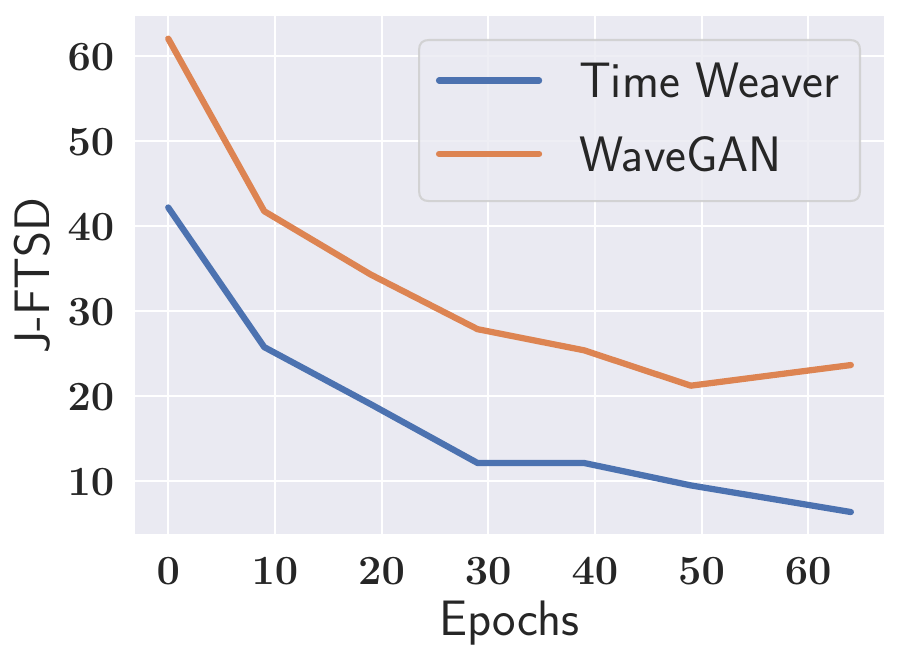}
\caption{\textbf{\timeweaver \space provides a better rate of decrease of \ac{jftsd} during training than \ac{gan}s.} Here, we show how \ac{jftsd} changes over epochs during the training phase for the Air Quality dataset. For this experiment, we compare \timeweaver-CSDI against WaveGAN. We observe that \timeweaver-CSDI provides a much higher rate of decrease of \ac{jftsd} when compared to WaveGAN, indicating better training.}
\label{fig:jftsd_training}
\end{figure}

In this section, we provide a quantitative analysis of the training and inference complexities of our proposed \timeweaver \space models. For training complexity, we show how the \ac{jftsd} metric varies over epochs during training in \cref{fig:jftsd_training}. We note that \timeweaver \space provides a faster rate of decrease of \ac{jftsd} when compared to \ac{gan}s.

For the inference complexity, we provide a theoretical discussion on the time complexity of \timeweaver \space in \cref{sec:experiments}. Additionally, we provide the measured inference latency for \timeweaver \space variants in \cref{tab:inference_latency}. From \cref{tab:inference_latency}, it can be seen that the inference latency for \timeweaver-CSDI is much lower than the inference latency for \timeweaver-SSSD for generating a single sample. However, we note that \timeweaver-SSSD is more suitable for batched generation as \timeweaver-CSDI drastically limits the maximum allowable batch size due to the forward pass through its feature transformer layers.

\begin{table}[h!]
\centering
\begin{sc}
\begin{tabular}{lllll}
\hline
                & & Inference Latency (s) & & \\ \hline
                 & Air Quality    & ECG             & Traffic        & Electricity    \\ \hline
\timeweaver-CSDI & 3.42$\pm$0.303 & 10.90$\pm$0.236 & 2.63$\pm$0.288 & 1.68$\pm$0.271 \\ \hline
\timeweaver-SSSD & 6.15$\pm$0.302   & 64.30$\pm$0.085 & 5.07$\pm$0.299 & 3.93$\pm$0.299 \\ \hline
\end{tabular}
\end{sc}
\caption{\textbf{Inference latency of \timeweaver \space models for generating one sample.} In this table, we show the inference latency in seconds for our proposed \timeweaver \space models for all datasets. We compute the mean inference latency and the standard deviation for generating one sample over ten runs. The inference experiments were performed on a single NVIDIA RTX A5000 GPU.}
\label{tab:inference_latency}
\end{table}
\subsection{Discussion On The Causal Relationship Between Time Series And Metadata}
\begin{figure}[H]
\centering
 \includegraphics[width=0.94\textwidth, keepaspectratio]{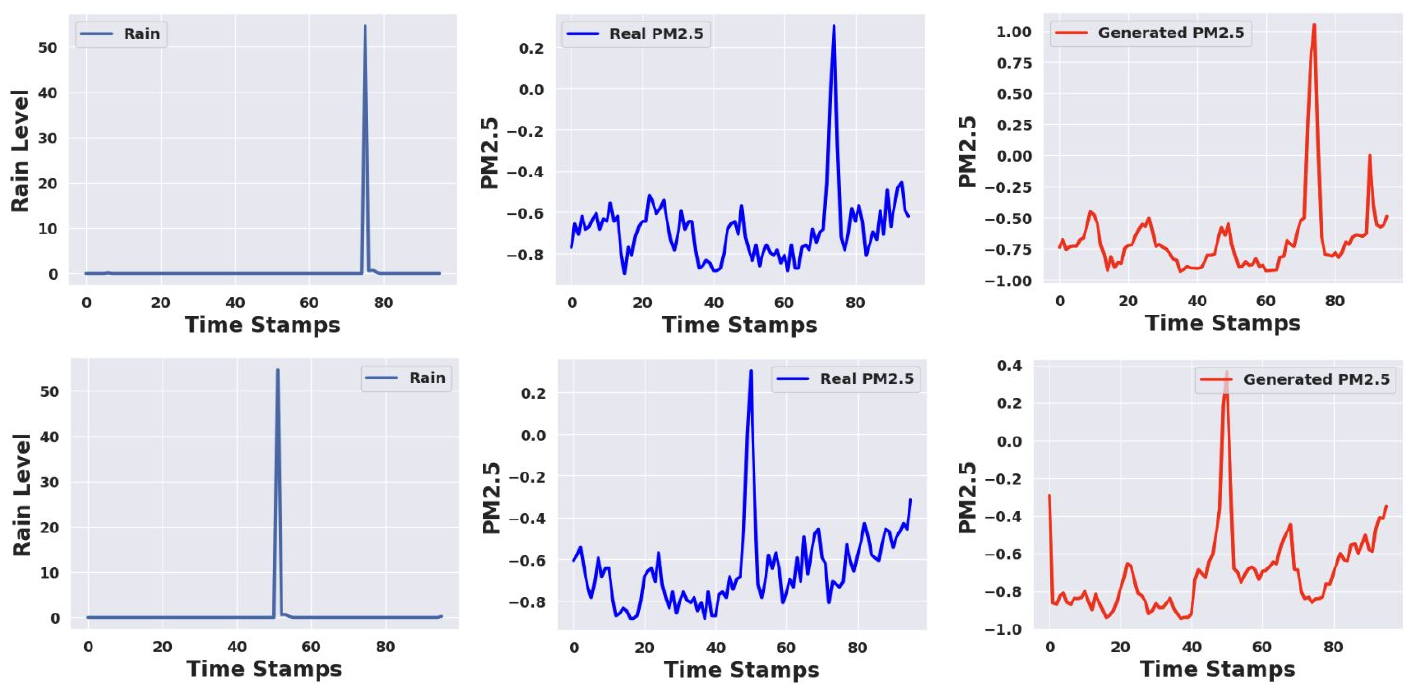}
\caption{\textbf{\timeweaver \space retains causal relationships between the input metadata and the generated time series.} In this figure, we showcase \timeweaver's ability to retain essential causal relationships between metadata and generated time series for the Air Quality dataset. In particular, we consider the effects of rainfall (metadata) on the particulate matter levels (PM2.5, one of the channels in the generated time series). We show two instances from the dataset with a sudden spike in rainfall. In both rows, the left image shows the rainfall level metadata and the center image shows the PM2.5 time series corresponding to the rainfall metadata from the dataset. The right image shows the generated PM2.5 time series channel for the rainfall metadata input. Observe that the sudden spike in the particulate matter level is faithfully replicated in the generated samples.}
\label{fig:causality}
\end{figure}

In this section, we present an interesting observation relevant to \timeweaver's ability to retain the causal effects of the input metadata on the generated time series. We consider the Air Quality dataset, where physical models exist that indicate the effect of rainfall (metadata) on the particulate matter levels (one of the time series channels).  We show that \timeweaver \space generates particulate matter levels that adhere to this effect through qualitative examples in \cref{fig:causality}. However, we note that this is not rigorous proof of causality. 

\subsection{Discussion On The Feature Extractors Used In The \ac{jftsd} Metric}
\begin{figure*}[!ht]
\centering
 \includegraphics[width=0.94\textwidth, keepaspectratio]{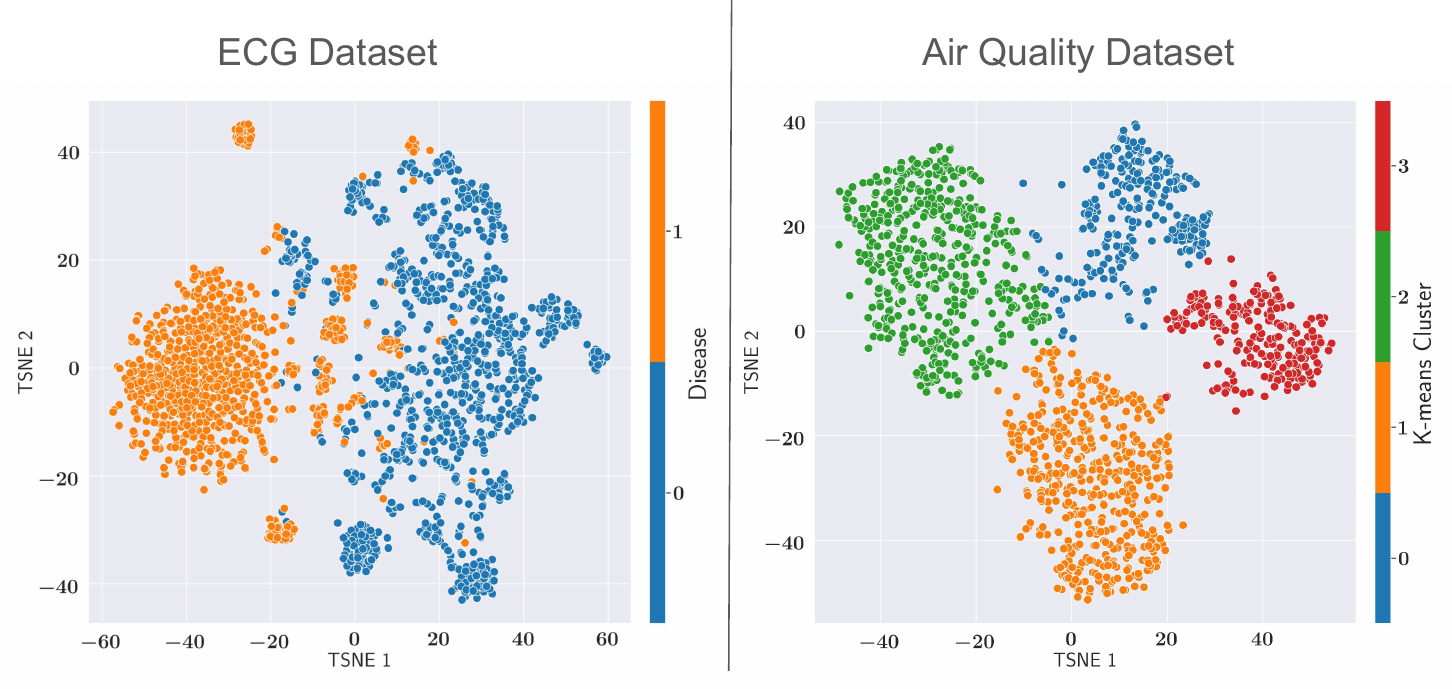}
\caption{\textbf{Interpretability of the time series embeddings obtained from \ac{jftsd}'s time series encoder.} In this figure, we show the clustering of time series embeddings obtained from the time series feature extractor $\metricembedderts$. We use t-SNE on the time series embeddings for dimensionality reduction. Note that the feature extractors ($\metricembedderts$ and $\metricembeddercondn$) are trained jointly using a contrastive learning-based approach (check Alg. 1). We show results for clustering on two datasets - ECG and Air Quality. For the ECG dataset, we showcase that the samples corresponding to the most common disease get clustered together. For the Air Quality dataset, we show that the time series embeddings get clustered into 4 clusters. However, due to the complexity of the metadata conditions (a combination of categorical and time-varying continuous conditions), we cannot interpret the clusters as in the case of the ECG dataset.}
\label{fig:tsne}
\end{figure*}

In this section, we analyze the embeddings from the feature extractors used to compute \ac{jftsd}. Note that these feature extractors ($\metricembedderts$ and $\metricembeddercondn$) are learned using a contrastive learning-based training approach similar to CLIP \cite{radford2021learning}. The feature extractors are jointly trained (check \cref{alg:metric_training}) to maximize the similarity between a time series embedding and its paired metadata embedding. This results in the clustering of time series embeddings for time series samples that have similar paired metadata. We test this observation on the ECG and the Air Quality datasets and show low-dimensional visualization of the time series embeddings in \cref{fig:tsne}. 

We use t-SNE on the time series embeddings for dimensionality reduction. For the ECG dataset, we observe that time series samples corresponding to the most common heart disease statement get clustered together. This is shown in \cref{fig:tsne}. Similarly, for the Air Quality dataset, we observe that the time series samples get clustered into 4 clusters. However, the dataset has a combination of categorical metadata conditions such as stations, years, months, etc, and continuous metadata conditions such as temperature, pressure, rain levels, etc. This complexity in metadata conditions inhibits us from interpreting the clusters, as in the case of the ECG dataset. Therefore, our critical insight is that the contrastive learning approach results in the clustering of the time series embeddings corresponding to similar paired metadata. However, the interpretability of the clusters is non-trivial for arbitrary combinations of categorical, continuous, and time-varying metadata conditions. 

\pagebreak
\subsection{Additional Qualitative Results} \label{appendix:additonal_qualitative}
In this section, we provide additional qualitative results generated using \timeweaver. 

\begin{figure}[H]
\centering
\includegraphics[width=1\textwidth]{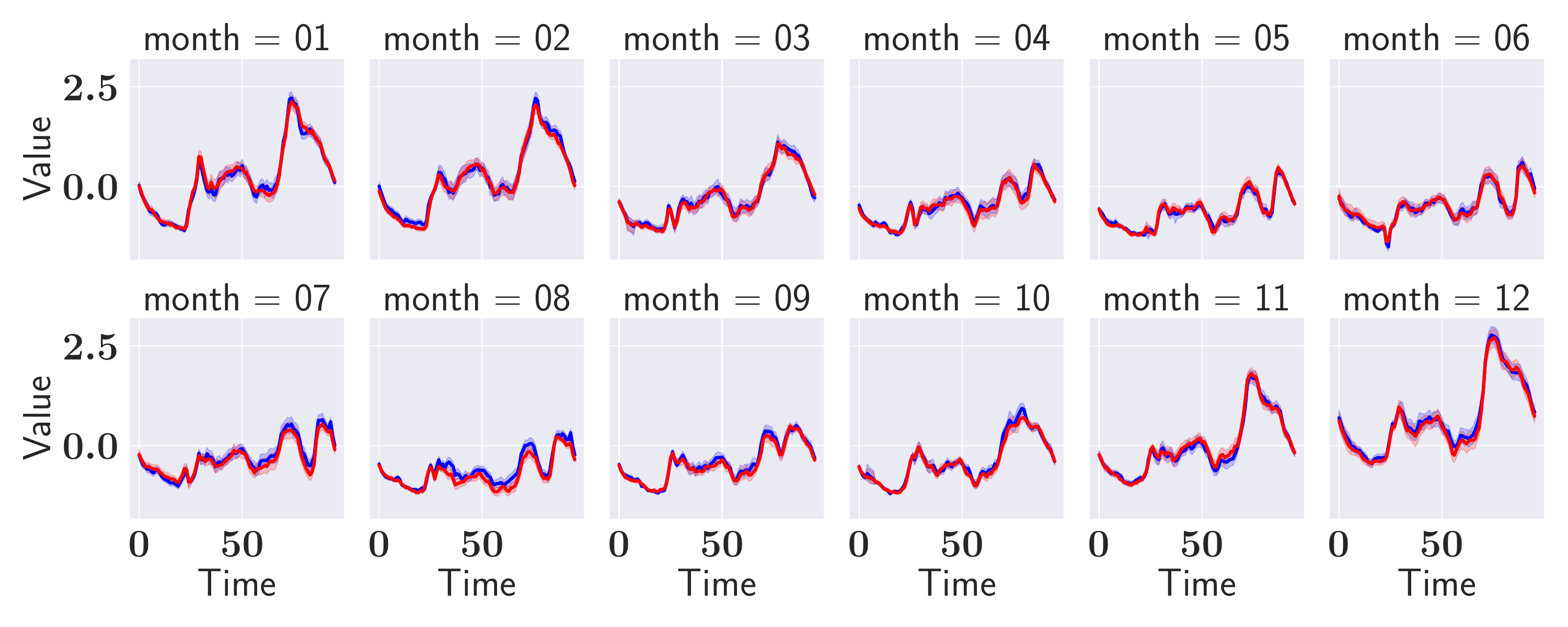}
\includegraphics[width=1\textwidth]{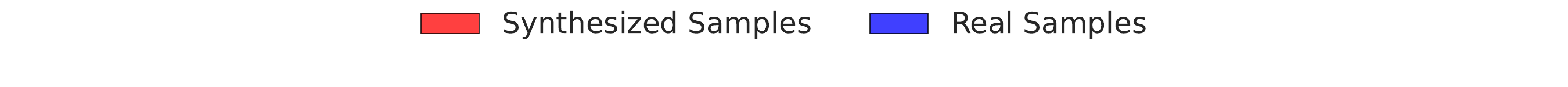}
\caption{\textbf{Generated time series samples from the Electricity dataset}}
\end{figure}
\begin{figure}[!ht]
\centering
\includegraphics[width=1.0\textwidth]{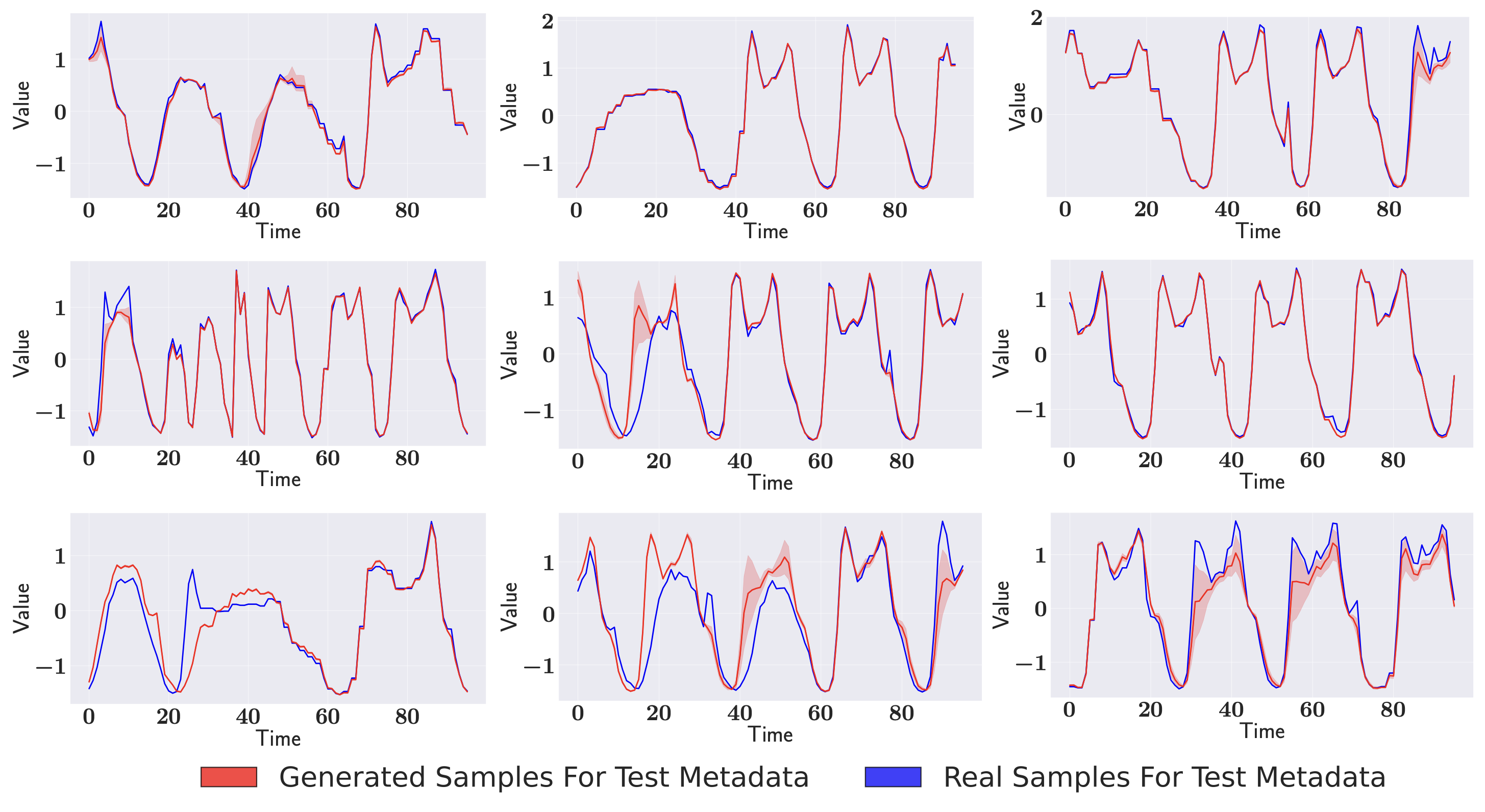}
\vspace{-0.5em}
\caption{\textbf{Qualitative Results from the \timeweaver-CSDI model for the Traffic dataset}   }
\vspace{-1em}
\label{fig:traffic_qualitative}
\end{figure}
\begin{figure}[ht]
\centering
\includegraphics[width=1.0\textwidth]{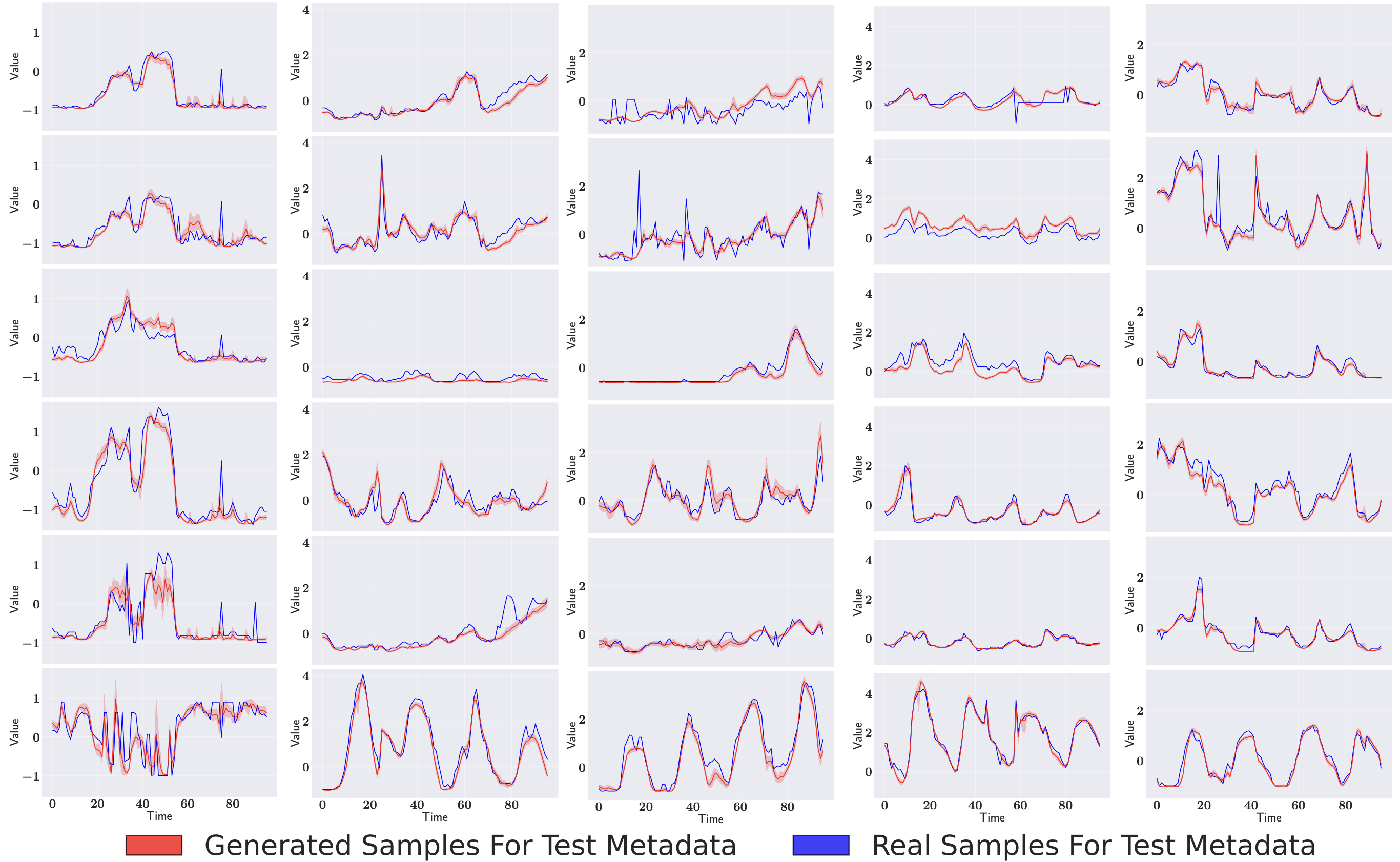}
\caption{\textbf{Qualitative results from the \timeweaver-CSDI model for the Air Quality dataset}   }
\label{fig:air_quality_qualitative}
\end{figure}

\begin{figure}[h]
\centering
\includegraphics[width=1.0\textwidth]{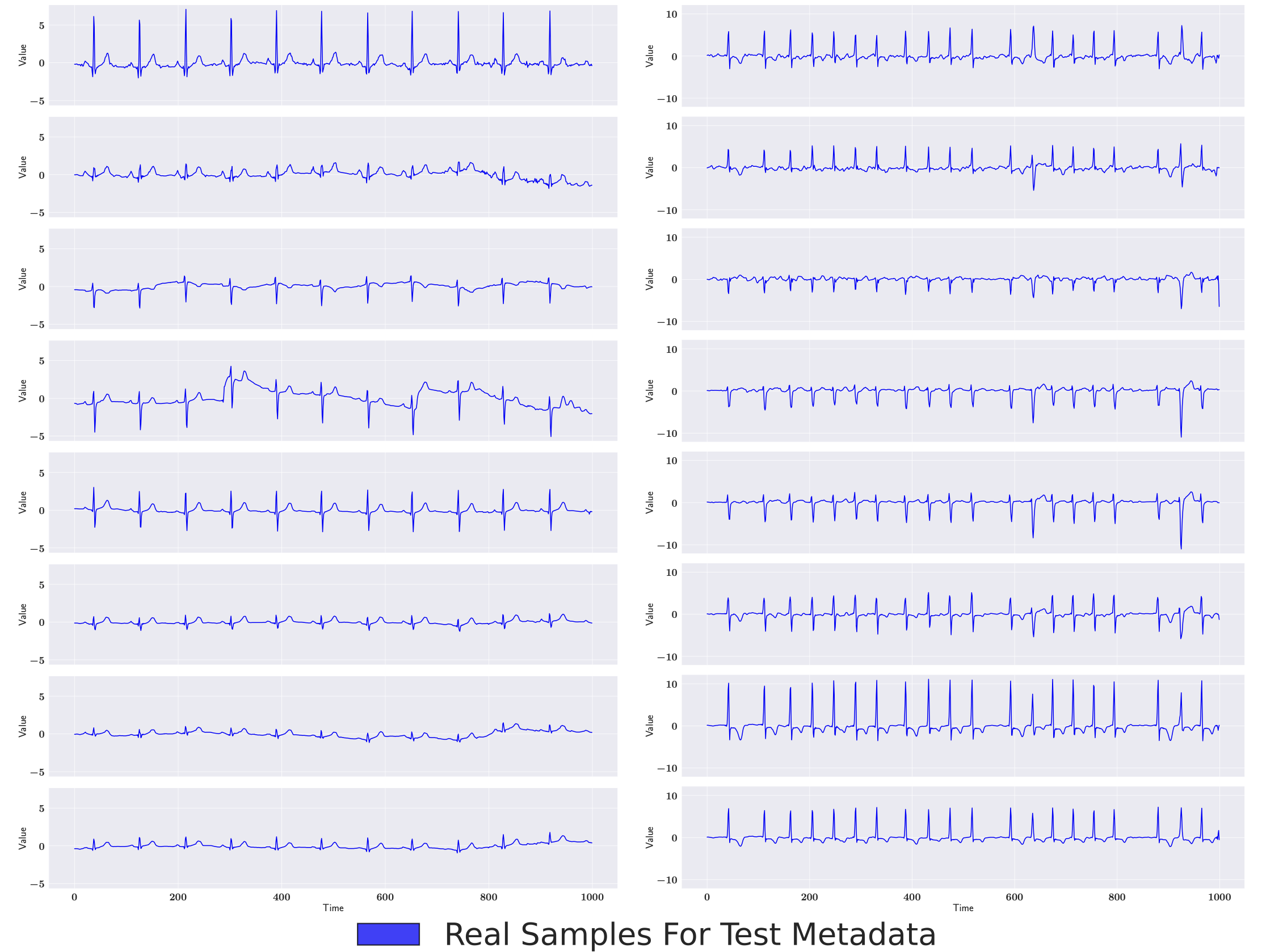}
\caption{\textbf{Real time series samples from the ECG dataset}}
\label{fig:ecg_real_qualitative}
\end{figure}
\begin{figure}[h]
\centering
\includegraphics[width=1.0\textwidth]{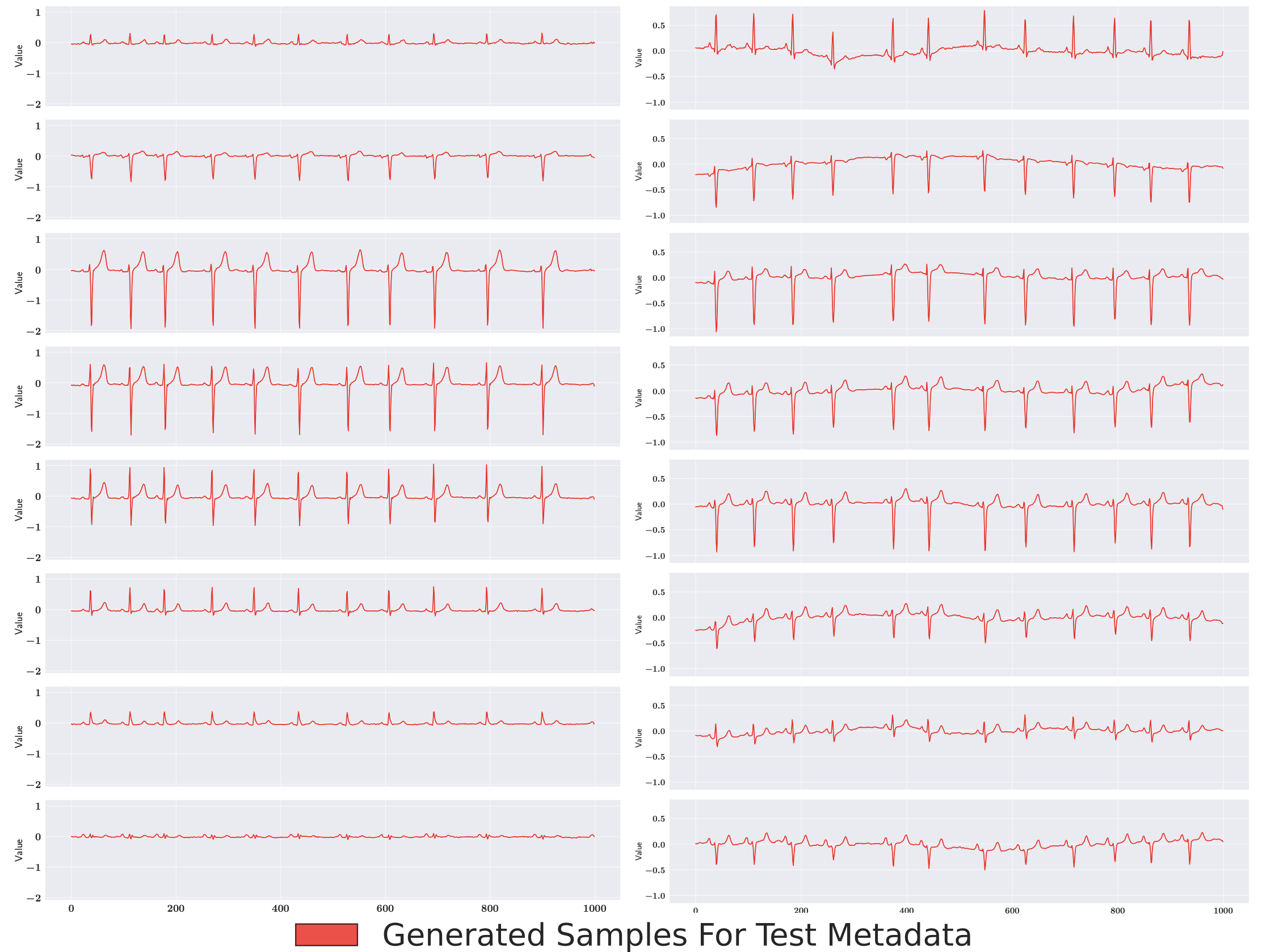}
\caption{\textbf{Generated time series samples from the ECG dataset}   }
\label{fig:ecg_generated_qualitative}
\end{figure}


\end{document}